%% file: main.tex
\pdfoutput=1
\documentclass[twoside,11pt]{article}

\usepackage{scrextend}
\usepackage{verbatim}
\usepackage{subcaption}
\usepackage{jmlr2e}
\usepackage{bm}
\usepackage{amsmath}
\usepackage{amssymb, amscd}
\usepackage{mathtools}
\usepackage{hhline}
\usepackage{multirow}
\usepackage{graphicx}
\usepackage{textcomp}
\usepackage{xcolor}
\usepackage{hyperref}
\hypersetup{
    colorlinks,
    linkcolor={red!50!black},
    citecolor={blue!50!black},
    urlcolor={blue!80!black}
}

\newcommand{\dataset}{{\cal D}}
\newcommand{\fracpartial}[2]{\frac{\partial #1}{\partial  #2}}
\newtagform{fn}{(}{)\footnotemark}

\ShortHeadings{Survey on the attention based RNN model and its applications in computer vision}{Feng Wang}
\firstpageno{1}

\makeatletter
\renewcommand\tableofcontents{%
  \null\hfill\textbf{\Large\contentsname}\hfill\null\par
  \@mkboth{\MakeUppercase\contentsname}{\MakeUppercase\contentsname}
  \@starttoc{toc}%
}
\makeatother

\begin{document}

\title{Survey on the attention based RNN model and its applications in computer vision}

\author{\name Feng Wang \email f.wang-6@student.tudelft.nl \\
       \addr Pattern Recognition Lab\\
       EEMCS\\
       Delft University of Technology\\
       Mekelweg 4, 2628 CD Delft, The Netherlands
       \AND
       \name D.M.J. Tax \email D.M.J.Tax@tudelft.nl \\
       \addr Pattern Recognition Lab\\
       EEMCS\\
       Delft University of Technology\\
       Mekelweg 4, 2628 CD Delft, The Netherlands}


\maketitle

\tableofcontents

\newpage

\input{abstract}

\input{introduction}

\input{attention-rnn}

\input{application}

\input{conclusion}

\newpage

\phantomsection
\addcontentsline{toc}{section}{References}
\bibliography{sample}

\end{document}

%% file: abstract.tex
\phantomsection
\addcontentsline{toc}{section}{Abstract}

\begin{abstract}
The recurrent neural networks (RNN) can be used to solve the sequence to sequence problem, where both the input and the output have sequential structures. Usually there are some implicit relations between the structures. However, it is hard for the common RNN model to fully explore the relations between the sequences. In this survey, we introduce some attention based RNN models which can focus on different parts of the input for each output item, in order to explore and take advantage of the implicit relations between the input and the output items. The different attention mechanisms are described in detail. We then introduce some applications in computer vision which apply the attention based RNN models. The superiority of the attention based RNN model is shown by the experimental results. At last some future research directions are given.
\end{abstract}


%% file: introduction.tex
\section{Introduction}

In this section, we discuss the nature of attention based recurrent neural network (RNN) model. What does "attention" mean? What are the applications of the attention based model in computer vision area? What is the attention based RNN model? What are the attention mechanisms introduced in this survey? What are the advantages of using attention based RNN model? These are the questions we would like to answer in this section.

\subsection{What does "attention" mean?}

In psychology, limited by the processing bottlenecks, humans tend to selectively concentrate on a part of the information, and at the same time ignore other perceivable information. The above mechanism is usually called attention \cite{anderson1990cognitive}. For example, in human visual processing, although human's eye has the ability to receive a large visual field, usually only a small part is fixated on. The reason is different areas of retina have different magnitude of processing ability, which is usually referred as acuity. And only a small area of the retina, fovea, has the greatest acuity. To allocate the limited visual processing resources, one needs to firstly choose a particular part of the visual field, and then focuses on it. For example, when humans are reading, usually the words to be read at the particular moment are attended and processed. As a result, there are two main aspects of attention:

\begin{itemize}
  \item Decide which part of the input needs to be focused on.
  \item Allocate the limited processing resources to the important part.
\end{itemize}

\begin{figure}
  \centering
  \includegraphics{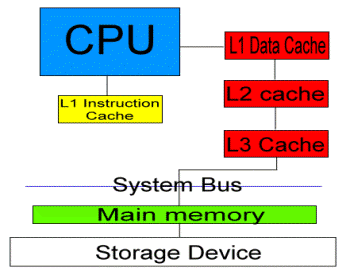}
  \caption{The CPU cache system. We can see that there is a multi-level structure between the data computing unit (CPU), and the data permanent storage device (Storage Device). The figure is taken from \cite{cache}.}
  \label{fig:intro-cache}
\end{figure}

The definition of attention introduced in psychology is very abstract and intuitive, so the idea is borrowed and widely used in computer science area, although some techniques do not contain exactly the word "attention". For example, a CPU will only load the data needed for computing instead of loading all data available, which can be seen as a na\"{\i}ve application of the attention. Furthermore, to make the data loading more effective, the \emph{multilevel storage} design in computer architecture is employed, i.e., the allocation of computing resources is aided by the structure of multi caches, main memory, and storage device as shown in Figure \ref{fig:intro-cache}. In a big picture, all of them make up an attention model.


\subsection{What are the applications of the attention based model in computer vision?}

As mentioned above, the attention mechanism plays an important role in human visual processing. Some researchers also bring the attention into computer vision area. As in the human perception, a computer vision system should also focus on the important part of the input image, instead of giving all pixels the same weights. A simple and widely used method is extracting local image features from the input. Usually the local image features can be points, lines, corners, or small image patches, and then some measurements are taken from the patches and converted into descriptors. Obviously, the distribution of local features' positions in a natural image is not uniform, which makes the system give different weights to different parts of the input image, and satisfies the definition of attention.

Saliency detection is another typical example directly motivated by human perception. As mentioned above, humans have ability to detect salient objects/regions rapidly and then attend on by moving the line of sight. This capability is also studied by the computer vision community, and many saliency object/region detection algorithms are proposed, e.g. \cite*{achanta2008salient, itti1998model, liu2011learning}. However, most of the saliency detection methods only use the low level image features, e.g., contrast, edge, intensity, etc., which makes them bottom-up, stimulus-driven, and fixed. So they cannot capture the task specific semantic information. Furthermore, the detections of different saliency objects/regions in an image are individually and independent, which is obviously unlike the humans, where the next attended region of an image usually is influenced by the previous perceptions.

The sliding window paradigm \cite*{dalal2005histograms, viola2001rapid} is another model which matches the essence of attention. It is common that the interested object only occupies a small region of the input image, and sometimes there are some limitations in the computational capabilities, so the sliding window is widely used in many computer vision tasks, e.g., object detection, object tracking, etc. However, since the size, shape and location of a window can be assigned arbitrarily, in theory there are infinite possible windows for an input image. Lots of work is devoted to reducing the number of windows to be evaluated, and some of them make notable speedups compared to the na\"{\i}ve method. But most of the window reducing algorithms are specifically designed for object detection or object tracking tasks, and the lack of universality makes them difficult to be applied in other tasks. Besides, most of the sliding window methods do not have the ability to take full advantage of the past processed data in future prediction.

\subsection{What is the attention based RNN model?}

Note that the \emph{attention} is just a mechanism, or a methodology, so there is no strict definition in mathematics what it is. For example, the local image features, saliency detection, sliding window methods introduced above, or some recently proposed object detection methods like \cite{gonzalez2015active}, all employ the attention mechanism in different mathematical forms. On the other hand, as illustrated later, the RNN is a specific type of neural network with a specific mathematical description. The \emph{attention based RNN models} in this survey refers in particular to the RNN models which are designed for sequence to sequence problems with the attention mechanism. Besides, all the systems in this survey are \emph{end-to-end} trainable, where the parameters for both the attention module and the common RNN model should be learned simultaneously. Here we will give a general explanation of what the RNN is and what the attention based RNN model is, and the mathematical details are left to be described later.

\subsubsection{Recurrent neural network (RNN)}

\begin{figure}
  \centering
  \makebox[\textwidth]{\includegraphics{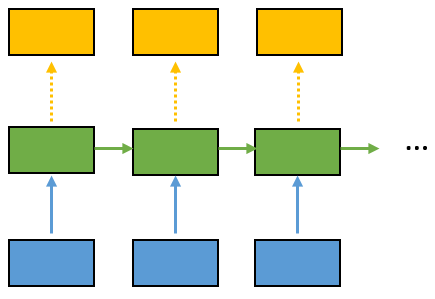}}
  \caption{An abstract structure of a RNN unit, where each rectangle represents a vector. The blue rectangles are the input vectors, the yellow rectangles are the output vectors, and the green rectangles are the hidden states. The dashed arrow indicates that the output is optional.}
  \label{fig:intro-rnn}
\end{figure}

Recently, the convolutional neural network (CNN) is very popular in the computer vision community. But the biggest problem for the CNN is that it only accepts a fixed length input vector and gives a fixed-length output vector. So it cannot handle the data with rich structures, especially the sequences. That is the reason why RNN is interesting, which can not only operate over sequences of input vectors, but also generate sequences of output vectors. Figure \ref{fig:intro-rnn} gives an abstract structure of a RNN unit.

The blue rectangles are the input vectors in a sequence, and the yellow rectangles are the output vectors in a sequence. By holding a hidden state (green rectangles in Figure \ref{fig:intro-rnn}), the RNN is able to process the sequence data. The dashed arrow indicates sometimes the output vectors do not have to be generated. The details of the structure of the neural network, the recurrent neural network, how the hidden state works, and how the parameters are learned will be described later.

\subsubsection{The RNN for sequence to sequence problem}

\paragraph{Sequence to sequence problem}

As mentioned above, the RNN unit holds a hidden state which empowers it to process sequential data with variable sizes. In this survey, we focus on the RNN models for sequence to sequence problem. The sequence to sequence problem is defined as to map the input sequence to the output sequence, which is a very general definition. Here the \emph{input sequence} does not refer strictly to a sequence of items, otherwise for different sequence to sequence problems, it could be in different forms. For example:

\begin{itemize}
  \item For the machine translation task, one usually needs to translate some sentences in the source language to the target language. In this case, the input sequence is a natural language sentence, where each item in the sequence is the word in the sentence. The order of the items in the sequence is critical.
  \item In the video classification task, a video clip usually should be assigned a label. In this case, the input sequence is a list of frames of the video, where each frame is an image. And the order is critical.
  \item The object detection task requires the model to detect some objects in an input image. In this case, the input sequence is only an image which can be seen as a list of objects. But obviously, it is not trivial to extract those objects from the image without additional data or information. So for the object detection task, the input of the model is just a single feature map (an image).
\end{itemize}

Similarly, the output sequence also does not always contain explicit items, but in this survey, we only focus on the problems where the output sequence has explicit items, although sometimes the number of items in the sequence is one. Besides, no matter the input is a feature map or it contains some explicit items, we always use "input sequence" to represent them. And the term "item" is used to describe the item in the sequence no matter it is contained by the sequence explicitly or implicitly.

\paragraph{The structure of the RNN model for sequence to sequence problem}

Under the condition that the lengths of the input and output sequences can vary, it is not possible for a common RNN to directly construct the corresponding relationships between the items in the input and output sequences without any additional information. As a result, the RNN model needs to firstly learn and absorb all (or a part of) useful information from all of the input items, and then make the predictions. So it is natural to divide the whole system into two parts: the encoder, and the decoder \cite{cho2015describing, SutskeverVL14}. The encoder encodes all (or a part of) the information of the input data to an intermediate code, and the decoder employs this code in generating the output sequence. In some sense, all neural networks can be cut into two parts in any position, the first half is called the encoder, and the second half is treated as the decoder, so here the encoder and decoder are both neural networks.

\begin{figure}
  \centering
  \makebox[\textwidth]{\includegraphics{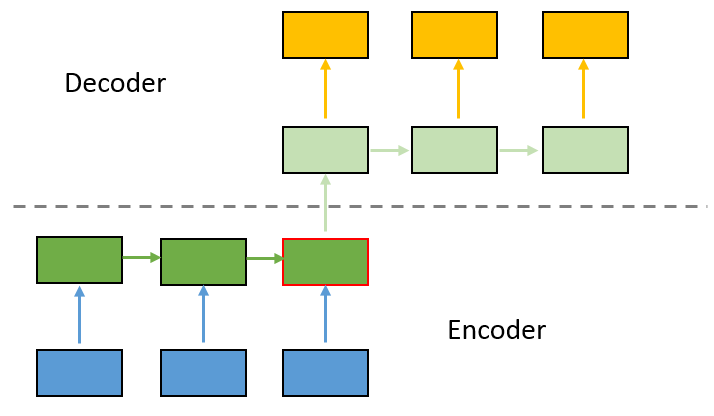}}
  \caption{The encoder and the decoder. The rectangle surrounded by the red box is the intermediate code. The blue rectangles represent the input items, the dark green rectangles are the hidden states of the encoder, the shallow green rectangles represent the hidden states of the decoder, and the yellow rectangles are the output items.}
  \label{fig:intro-en-de-2}
\end{figure}

Figure \ref{fig:intro-en-de-2} gives a visual illustration of a RNN model for sequence to sequence problem. The blue rectangles represent the input items, the dark green rectangles are the hidden states of the encoder, the shallow green rectangles represent the hidden states of the decoder, and the yellow rectangles are the output items. Although there are three input items and three output items in Figure \ref{fig:intro-en-de-2}, actually the number of input and output items can be arbitrary number and do not have to be the same. And the rectangle surrounded by the red box in Figure \ref{fig:intro-en-de-2} serves as the "middleman" between the input and the output which is the intermediate code generated by the encoder. In Figure \ref{fig:intro-en-de-2} both the encoder and the decoder are recurrent networks, while with different types of the input sequence mentioned above, the encoder does not have to be recurrent all the time. For example, when the input is a feature map, a neural network without recurrence usually is applied as the encoder.

When dealing with a supervised learning problem, during training, usually the ground truth output sequence corresponds to an input sequence is available. With the predicted output sequence, one can calculate the log-likelihood between the items which have the same indices in the predicted and the ground truth output sequences. The sum of the log-likelihood of all items usually is set as the objective function of the supervised learning problem. And then usually the gradient decent/ascent is used to optimize the objective in order to learn the value of the parameters of the model.

\subsubsection{Attention based RNN model}

For sequence to sequence problems, usually there are some corresponding relations between the items in the output and input sequences. However for a common RNN model introduced above, the length of the intermediate code is fixed, which prevents the model giving different weights to different items in an input sequence explicitly, so all items in the input sequence has the same importance no matter which output item is attempted to be predicted. This inspires the researchers to add the attention mechanism into the common RNN model.

Besides, the encoding-decoding process can also be regarded as a compression process: firstly the input sequence is compressed into an intermediate code, and then the decoder decompresses the code to get the output. The biggest problem is that no matter what kind of input the model gets, it always compresses the information into a fixed length code, and then the decoder uses this code to decompress all items of a sequence. From an information theory perspective, this is not a good design since the length of the compressed code should have a linear relationship to the amount of information the inputs contain. As in the real file compression, the length of the compressed file is proportional to the amount of information the source files contain. There are some straightforward solutions to solve the problem mentioned above:

\begin{enumerate}
  \item Build an extremely non-linear and complicated encoder model such that it can store large amount of information into the intermediate code.
  \item Just use a long enough code.
  \item Stay with the fixed length code, but let the encoder only encode a part of the input items which are needed for the current decoding.
\end{enumerate}

Because the input sequence can be infinite long (or contain infinite amount of information), the first solution does not solve the problem essentially. Needless to say, it is even a more difficult task to estimate the amount of information some data contains. Based on the same reason, solution 2 is not a good choice either. In some extreme cases, even if the amount of information is finite, the code should still be large enough to store all possible input-output pairs, which is obviously not practical.

Solution 3 also cannot solve the problem when the amount of the input information needed for the current decoding is infinite. But usually a specific item in the output sequence only corresponds to a part of the input items. In this case, solution 3 is more practical and can prevent the problem in some degrees compared to other solutions if the needed input items can be "located" correctly. Furthermore, it essentially has the same insights as the multilevel storage design pattern as shown in Figure \ref{fig:intro-cache}. The RNN model which has the addressing ability is called the attention based RNN model.

With the attention mechanism, the RNN model is able to assign different weights to different parts of items in the input sequence, and consequently, the inherent corresponding relations between items in input and output sequences can be captured and exploited explicitly. Usually, the attention module is an additional neural network which is connected to the original RNN model whose details will be introduced later.

\subsection{What are the attention mechanisms introduced in this survey?}

In this survey, we introduce four types of attention mechanisms, which are: item-wise soft attention, item-wise hard attention, location-wise hard attention, and location-wise soft attention. Here we only give a brief and high-level illustration of them, and leave the mathematical descriptions in the next chapter.

For the \emph{item-wise} and the \emph{location-wise}, as mentioned above, the forms of the input sequence are different for different tasks. The item-wise mechanism requires that the input is a sequence containing explicit items, or one has to add an additional pre-processing step to generate a sequence of items from the input. However the location-wise mechanism is designed for the input which is a single feature map. The term "location" is used because in most cases this kind of input is an image, and when treating it as a sequence of objects, all the objects can be pointed by their locations.

As described above, the item-wise method works on "item-level". After being fed into the encoder, each item in the input sequence has an individual code, and all of them make up a code set. During decoding, at each step the item-wise soft attention just calculates a weight for each code in the code set, and then makes a linear combination of them. The combined code is treated as the input of the decoder to make the current prediction. The only difference lies in the item-wise hard attention is that it instead makes a hard decision and stochastically picks some (usually one) codes from the code set according to their weights as the intermediate code fed into the decoder.

The location-wise attention mechanism directly operates on an entire feature map. At each decoding step, the location-wise hard attention mechanism discretely picks a sub-region from the input feature map and feeds it to the encoder to generate the intermediate code. And the location of the sub-region to be picked is calculated by the attention module. The location-wise soft attention still accepts the entire feature map as input at each step while otherwise makes a transformation on it in order to highlight the interesting parts instead of discretely picking a sub-region.

As mentioned above, usually the attention mechanism is implemented as an additional neural network connected to the raw RNN. The whole model should still be \emph{end-to-end}, where both the raw RNN and the attention module are learned simultaneously. When it comes to \emph{soft} and \emph{hard}, for the soft attention, the attention module is differentiable with respect to the inputs, so the whole system can still be updated by gradient ascent/decent. However, since the hard attention mechanism makes hard decisions and gives discrete selections of the intermediate code, the whole system is not differentiable with respect to its inputs anymore. Then some techniques from the reinforcement learning are used to solve learning problem.

\begin{figure}
  \centering
  \makebox[\textwidth]{\includegraphics[width=\textwidth]{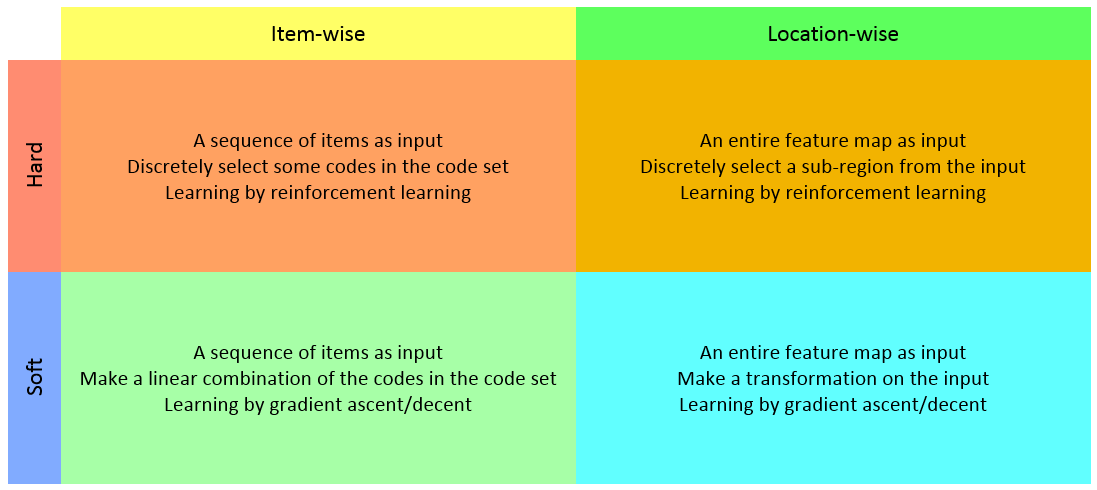}}
  \caption{Four attention mechanisms.}
  \label{fig:intro-attentions}
\end{figure}

In summary, the attention mechanisms are designed to help the model select better intermediate codes for the decoder. Figure \ref{fig:intro-attentions} gives a visual illustration of the four attention mechanisms which are about to be introduced in this survey.

\subsection{What are the advantages of using attention based RNN model?} \label{sec:intro-adv}

Here we give some general advantages of the attention based RNN model. The detailed comparison between the attention based RNN model and the common RNN model will be illustrated later. Advantages:

\begin{itemize}
  \item First of all, as its name indicates, the attention based RNN model is able to learn to assign weights to different parts of the input instead of treating all input items equally, which can provide the inherent relations between the items the in input and output sequence. This is not only a way to boost the performance of some tasks, but it is also a powerful tool of visualization compared to the common RNN model.
  \item The hard attention model does not need to process all items in the input sequence, instead, sometimes only processes the interested ones, so it is very useful for some tasks with only partially observable environments, like game playing, which usually cannot be handled by the common RNN model.
  \item Not limited in computer vision area, the attention based RNN model is also suitable for all kinds of sequence related problems. For example:
    \begin{itemize}
      \item Applications in natural language processing (NLP), e.g., machine translation \cite{BahdanauCB14, MengLWLJL15}, machine comprehension \cite{HermannKGEKSB15}, sentence summarization \cite{RushCW15}, word representation \cite{LingDBT15, LingTAFDBTL15}.
      \item Bioinformatics \cite{SonderbySNW15}.
      \item Speech recognition \cite{ChanJLV15, MeiBW15}.
      \item Game play \cite{mnih-atari-2013}.
      \item Robotics.
    \end{itemize}
\end{itemize}

\subsection{Organization of this survey}

In Section \ref{sec:att}, we describe the general mathematical form of the attention based RNN model, especially four types of the attention mechanisms: item-wise soft attention, item-wise hard attention, location-wise hard attention, and location-wise soft attention. Next, we introduce some typical applications of the attention based RNN model in computer vision area in Section \ref{sec:app}. At last, we summarize the insights of the attention based RNN model, and discuss some potential challenges and future research directions.

%% file: attention-rnn.tex
\section{The attention based RNN model} \label{sec:att}

One can have a general idea of what the attention based RNN is from the last section. In this section, we firstly give a short introduction of the neural network, followed by an abstract mathematical description of the RNN. And then four attention mechanisms on RNN model are analyzed in detail.

\subsection{Neural network}

\begin{figure}
  \makebox[\textwidth]{\includegraphics{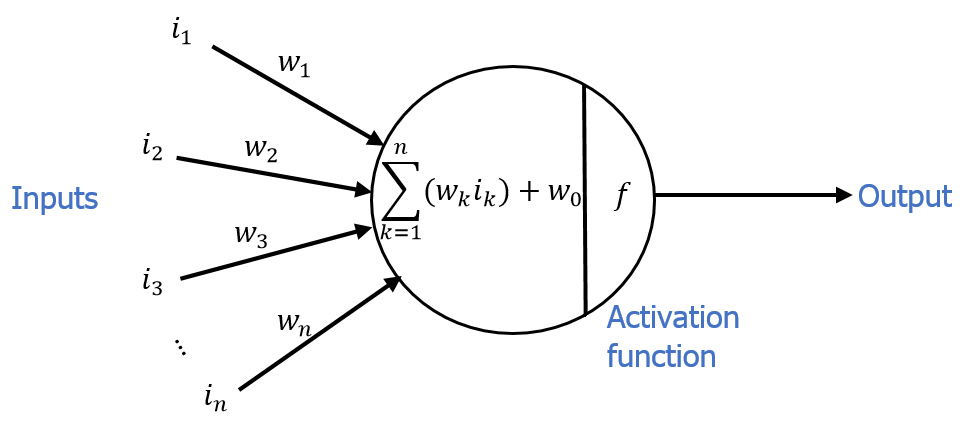}}
  \caption{A neuron.}
  \label{fig:att-neuron}
\end{figure}

A neural network is a data processing and computational model consists of multiple neurons which connect to each other. The basic component of all neural network is the neuron whose name indicates its inspiration from the human neuron cell. Ignoring the biological implication, a neuron in the neural network accepts one or more inputs, reweighs the inputs and sums them, and finally gives one output by passing the sum through an activation function. Let the input to the neuron be $\pmb{i}=<i_1,i_2,\dots,i_n>$, the corresponding weights be $\pmb{w}=<w_1,w_2,\dots,w_n>$, and the output be $o$. A visual example of a neuron is given in Figure \ref{fig:att-neuron}, where $f$ is the activation function. And we have:

\begin{equation}\label{eq:att-neuron}
  o = f \left( \sum_{k=1}^{n} (w_k i_k) + w_0 \right)
\end{equation}

where $w_0$ is the bias parameter. If we add $i_0=1$ into $\pmb{i}$ and rewrite the previous equation in vector form, then

\begin{equation}\label{eq:att-neuron-vec}
  o = f \left( \sum_{k=0}^{n} (w_k i_k) \right) = f\left( \pmb{w}^T\pmb{i} \right)
\end{equation}

In summary, a neuron just applies an activation function on the linear transformation of the input parameterized by the weights. Usually the activation functions are designed to be non-linear in order to import non-linearity into the model. There are many different forms of activation functions, i.e., sigmoid, tanh, ReLU \cite{NairH10}, PReLU \cite{HeZR015}.

\begin{figure}
  \makebox[\textwidth]{\includegraphics{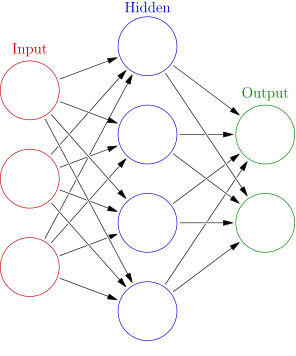}}
  \caption{A neural network, where each circle represents a neuron, and each arrow represents a connection. Here the "connection" means the output of a neuron is used as the input for another neuron. The figure is taken from \cite{NN}.}
  \label{fig:att-NN}
\end{figure}

A neural network is usually constructed by the connections of many neurons, and the most popular structure is layer by layer connection as shown in Figure \ref{fig:att-NN}. In Figure \ref{fig:att-NN}, each circle represents a neuron, and each arrow represents a connection between neurons. The outputs of all neurons in a layer in Figure \ref{fig:att-NN} are connected to the next layer, which can be categorized as fully-connected layer. In summary, if we treat the neural network as a black box and see it from outside, a simple mathematical form can be obtained:

\begin{equation}\label{eq:att-NN}
  \pmb{o} = \phi_{W} (\pmb{i})
\end{equation}

where $\pmb{i}$ is the input, $\pmb{o}$ is the output, and $\phi_W$ is a particular model consists of many neurons parameterized by $W$. One property of the model $\phi_W$ is that it is usually differentiable with respect to $W$ and $\pmb{i}$. For more details about the neural network, one can refer to \cite{Schmidhuber15}.

The convolutional neural network (CNN) is a specific type of neural network with structures designed for image inputs \cite{lecun1998gradient}, which usually consists of multiple convolutional layers followed by a few fully-connected layers. The convolutional layer can take a region of a 2D feature map as input, and this is the reason why it is suitable for image data. Recently the CNN is one of the most popular model as candidate for image related problems. A deep CNN model trained on some large image classification dataset (e.g., ImageNet \cite{DengDSLL009}) has good generalization power/ability and can be easily transferred to other tasks. For more details about CNN, one can refer to \cite{Schmidhuber15}.

\subsection{A general RNN unit}

As its name shows, the RNN is also a type of neural network, which therefore consists of a certain number of neurons as all kinds of neural network. The "recurrent" in the RNN indicates that it performs the same operations for each item in the input sequence, and at the same time, keeps a \emph{memory} of the processed items for future operations. Let the input sequence be $I=\left( \pmb{i}_1, \pmb{i}_2, \dots, \pmb{i}_T \right)$. At each step $t$, the operation performed by a general RNN unit can be described by:

\begin{equation}\label{eq:att-rnn}
  \begin{bmatrix}
\pmb{\hat{o}}_t \\
\pmb{h}_t \\ \end{bmatrix} = \phi_{W}\left( \pmb{i}_{t}, \pmb{h}_{t-1} \right)
\end{equation}

where $\pmb{\hat{o}}_t$ is the predicted output vector at time $t$, and $\pmb{h}_t$ is the hidden state at time $t$. $\phi_W$ is a neural network parameterized by $W$ which takes the $t$-th input item ($\pmb{i}_t$), and the previous hidden state $\pmb{h}_{t-1}$ as inputs. Here it is clear that the hidden state $\pmb{h}$ performs as a memory to store the previous processed information. From Equation \eqref{eq:att-rnn}, we can see the definition of a RNN unit is quite general, because there is no specific requirements for the structure of the network $\phi_W$. People have already proposed hundreds or even thousands of different structures which are out of the scope of this paper. Now two widely used structures are LSTM (long-short term memory) \cite{HochreiterS97} and GRU (gated recurrent units) \cite{ChoMBB14}.

\subsection{The model for sequence to sequence problem}

As shown in Figure \ref{fig:intro-en-de-2}, the model for sequence to sequence problem can be generally divided into two parts: the encoder, and the decoder, where both of them are neural networks. Let the input sequence be $X$, and the output sequence be $Y$. As mentioned above, in some tasks the input sequence does not consists of explicit items, for which we use $X$ to represent the input sequence, and otherwise $X=\left(\pmb{x}_1,\pmb{x}_2,\dots,\pmb{x}_T \right)$. Besides, in this survey, only the output sequence containing explicit items is considered, i.e., $Y=\left(\pmb{y}_1,\pmb{y}_2,\dots,\pmb{y}_{T'} \right)$ is the output sequence. The lengths of the input and the output sequences do not have to be the same.

\subsubsection{Encoder}

As mentioned above, the core task of an encoder is to encode all (or a part of) the input items to an intermediate code to be decoded by the decoder. In the common RNN for sequence to sequence problem, the model does not try to figure out the corresponding relations between the items in the input and output sequences, and usually all items in the input sequence are compressed into a single intermediate code. So for the encoder,

\begin{equation}\label{eq:att-en}
  \pmb{c} = \phi_{W_{enc}}\left(X \right)
\end{equation}

where $\phi_{W_{enc}}$ represents the encoder neural network parameterized by $W_{enc}$, $\pmb{c}$ is the intermediate code to be used by the decoder. Here the encoder can be any neural network, but its type usually depends on the input data. For example, when the input $X$ is treated as a single feature map, usually a neural network without recurrence is used as the encoder, like a CNN for the image input. But it is also common to set encoder as a recurrent network when the input is a sequence of data, e.g., a video clip, a natural language sentence, a human speech clip.

\subsubsection{Decoder} \label{sec:att-s2s-de}

If the length of the output sequence is larger than 1, in most cases the decoder is recurrent, because the decoder at least needs to have the knowledge what it has already predicted to prevent the repeated prediction. Especially in the RNN model with the attention mechanism which will be introduced later, the weights of the input items are assigned with the guidance of the past predictions, so in this case, the decoder should be able to store some history information. As a result, in this survey we only consider the cases where the decoder is an RNN.

As mentioned above, in a common RNN, the decoder accepts the intermediate code $\pmb{c}$ as input, and at each step $j$ generates the predicted output $\pmb{\hat{y}}_j$, and the hidden state $\pmb{h}_j$ by

\begin{equation}\label{eq:att-de}
  \begin{bmatrix}
\pmb{\hat{y}}_j \\
\pmb{h}_j \\ \end{bmatrix} = \phi_{W_{dec}}\left(\pmb{c}, \pmb{\hat{y}}_{j-1}, \pmb{h}_{j-1} \right)
\end{equation}

where $\phi_{W_{dec}}$ represents the decoder recurrent neural network parameterized by $W_{dec}$, which accepts the code $\pmb{c}$, the previous hidden state $\pmb{h}_{j-1}$, and the previous predicted output $\pmb{\hat{y}}_{j-1}$ as input. After $T'$ steps of decoding, a sequence of predicted outputs $\hat{Y} = \left( \pmb{\hat{y}}_{1},\pmb{\hat{y}}_{2},\dots,\pmb{\hat{y}}_{T'} \right)$ is obtained.

\subsubsection{Learning}

Like many other supervised learning problems, the supervised sequence to sequence problem is also optimized by maximizing the log-likelihood. With the input sequence $X$, the ground truth output sequence $Y$, and the predicted output sequence $\hat{Y}$, the log-likelihood for the $j$-th item $\pmb{y}_j$ in the output sequence $Y$ is

\begin{equation}\label{eq:att-ll}
  L_j \left( X, Y, \theta \right) = \log p\left(\pmb{y}_j|X, \pmb{y}_1, \pmb{y}_2, \dots, \pmb{y}_{j-1}, \theta \right) = \log p\left(\pmb{y}_j| \pmb{\hat{y}}_j, \theta \right)
\end{equation}

where $\theta=\left[W_{enc},W_{dec} \right]$ represents all learnable parameters in the model, and $p()$ calculates the likelihood of $\pmb{y}_j$ based on $\pmb{\hat{y}}_j$. For example, in a classification problem where $\pmb{y}_j$ indicates the index of the label, the $p\left(\pmb{y}_j| \pmb{\hat{y}}_j, \theta \right) = \mbox{softmax}(\pmb{\hat{y}}_j)_{\pmb{y}_j} $. The objective function is then set as the sum of the log-likelihood:

\begin{equation}\label{eq:att-sum-ll}
  L(X, Y, \theta) = \sum_{j=1}^{T'} L_j (X, Y, \theta) = \sum_{j=1}^{T'} \log p\left(\pmb{y}_j|X, \pmb{y}_1, \pmb{y}_2, \dots, \pmb{y}_{j-1}, \theta \right) = \sum_{j=1}^{T'} \log p\left(\pmb{y}_j| \pmb{\hat{y}}_j, \theta \right)
\end{equation}

For the common encoder-decoder system, both the encoder and decoder network are differentiable with respect to their parameters, $W_{enc}$, and $W_{dec}$, respectively, so the objective function is differentiable with respect to $\theta$. As a result, one can update the parameters $\theta$ by gradient ascent to maximize the sum of the log-likelihood.

\subsection{Attention based RNN model}

In this section, we give a comprehensive mathematical description of the four attention mechanisms. As mentioned above, essentially the attention module helps the encoder calculate a better intermediate code $\pmb{c}$ for the decoder. So in this section, the decoder part is identical as shown in Section \ref{sec:att-s2s-de}, but the encoder network $\phi_{W_{enc}}$ may not always be the same as in Equation \eqref{eq:att-en}. In order to make the illustration clear, we still keep the term $\phi_{W_{enc}}$ to represent the encoder.

\subsubsection{Item-wise soft attention}

The item-wise soft attention requires the input sequence $X$ contains some explicit items $\pmb{x}_1,\pmb{x}_2,\dots,\pmb{x}_T$. Instead of extracting only one code $\pmb{c}$ from the input $X$, the encoder of the item-wise soft attention RNN model extracts a set of codes $C$ from $X$:

\begin{equation}\label{eq:att-att-isa-C}
  C = \left\{ \pmb{c}_1, \pmb{c}_2, \dots, \pmb{c}_{T''} \right\}
\end{equation}

where the size of $C$ does not have to be the same as $X$, which means one can use multiple input items to calculate a single code or extract multiple codes from one input item. For simplicity, we just set $T''=T$. So

\begin{equation}\label{eq:att-att-isa-cs}
  \pmb{c}_t = \phi_{W_{enc}} \left( \pmb{x}_t \right) \mbox{ for } t \in \left(1,2,\dots,T \right)
\end{equation}

where $\phi_{W_{enc}}$ is the encoder neural network parameterized by $W_{enc}$. Note here that the encoder is different from the encoder in Equation \eqref{eq:att-en}, because the encoder in the item-wise soft attention model only takes one item of the sequence as input.

During the decoding, the intermediate code $\pmb{c}$ fed into the decoder is calculated by the attention module, which accepts all individual codes $C$, and the decoder's previous hidden state $\pmb{h}_{j-1}$ as input. At the decoding step $j$, the intermediate code is:

\usetagform{fn}
\begin{equation}\label{eq:att-att-isa-c}
  \pmb{c} = \pmb{c}^j = \phi_{W_{att}} \left( C, \pmb{h}_{j-1} \right)
\end{equation}

\footnotetext{For simplicity, in the following sections, we just use $\pmb{c}$ to represents the intermediate code fed into the decoder, and ignore the superscript $j$.}

\usetagform{default}

where $\phi_{W_{att}}$ represents the attention module parameterized by $W_{att}$. Actually, as long as the attention module $\phi_{W_{att}}$ is differentiable with respect to $W_{att}$, it satisfies the requirements of the soft attention. Here we only introduce the first proposed item-wise soft attention model in \cite{BahdanauCB14} for natural language processing. In this item-wise soft attention model, at decoding step $j$, for each input code $\pmb{c}_t$, a weight $\alpha_{tj}$ is calculated by:

\begin{equation}\label{eq:att-att-isa-e}
  e_{jt} = f_{att} \left( \pmb{c}_t, \pmb{h}_{j-1} \right)
\end{equation}

and

\begin{equation}\label{eq:att-att-isa-alpha}
  \alpha_{jt} = \frac {\exp(e_{jt})} {\sum_{t=1}^{T} \exp(e_{jt}) }
\end{equation}

where $f_{att}$ usually is also a neural network within the attention module and calculates the unnormalized weight $e_{jt}$. Here the normalized weight $\alpha_{tj}$ is explained as the probability that how the code $\pmb{c}_t$ is relevant to the output $\pmb{y}_j$, or the importance should be assigned to the $t$-th input item when making the $j$-th prediction. Note here that Equation \eqref{eq:att-att-isa-alpha} is a simple softmax function transforming the scores $e_{jt}$ to the scale from 0 to 1, which makes the $\alpha_{jt}$ can be interpreted as a probability.

Since now we have the probabilities, then the code $\pmb{c}$ is just calculated by taken the expectation of all $\pmb{c}_t$s with their probabilities $\alpha_{jt}$s:

\begin{equation}\label{eq:att-att-isa-finalc}
  \begin{aligned}
\pmb{c} &= \phi_{W_{att}} \left( C, \pmb{h}_{j-1} \right) \\
&= \mathop{\mathbb{E}}(\pmb{c}_t) \\
&= \sum_{t=1}^{T} \alpha_{jt} \pmb{c}_t
\end{aligned}
\end{equation}

\begin{figure}[!ht]
  \centering
  \makebox[\textwidth]{\includegraphics[width=\textwidth]{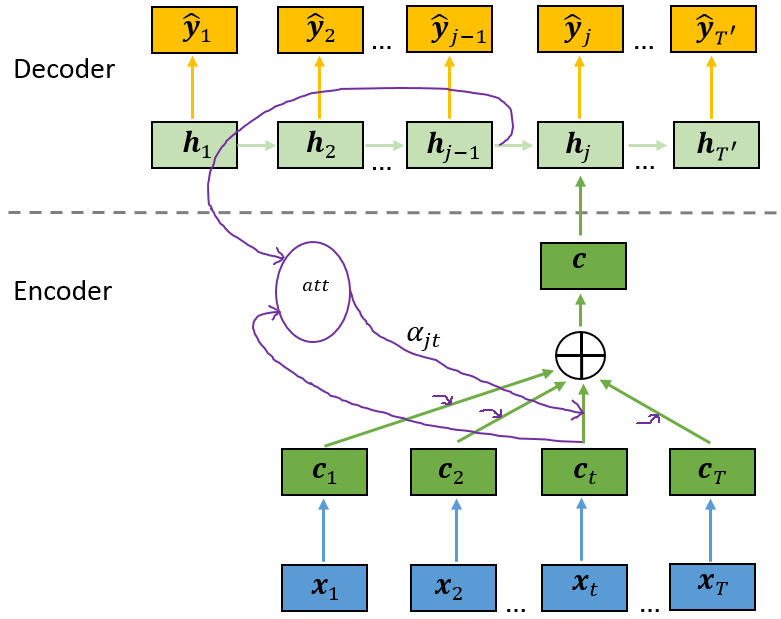}}
  \caption{The item-wise soft attention based RNN model at decoding step $j$. The purple ellipse represents the Equation \eqref{eq:att-att-isa-e} and Equation \eqref{eq:att-att-isa-alpha}. And the encoder is not recurrent.}
  \label{fig:att-att-isa-ex1}
\end{figure}

Figure \ref{fig:att-att-isa-ex1} gives a visual illustration of how the item-wise soft attention works at decoding step $j$, where the purple ellipse represents the Equation \eqref{eq:att-att-isa-e} and Equation \eqref{eq:att-att-isa-alpha}, and the encoder network is not recurrent.

\subsubsection{Item-wise hard attention}\label{sec:att-att-iha}

The item-wise hard attention is very similar to the item-wise soft attention. It still needs to calculate the weights for each code as shown from Equation \eqref{eq:att-att-isa-C} to Equation \eqref{eq:att-att-isa-alpha}. As mentioned above, the $\alpha_{jt}$ can be interpreted as the probability that how the code $\pmb{c}_t$ is relevant to the output $\pmb{y}_j$, so instead of a linear combination of all codes in $C$, the item-wise hard attention stochastically picks one code based on their probabilities. In detail, an indicator $l_j$ is generated from a categorical distribution at decoding step $j$ to indicate which code should be picked:

\begin{equation}\label{eq:att-att-iha-l}
  l_j \sim \mathcal{C}\left(T, \left\{ \alpha_{jt} \right\}_{t=1}^T\right)
\end{equation}

where $\mathcal{C}()$ is a categorical distribution parameterized by the probabilities of the codes ($\left\{\alpha_{jt} \right\}_{t=1}^T $). And $l_j$ works as an index in this case:

\begin{equation}\label{eq:att-att-iha-pick}
  \pmb{c} = \pmb{c}_{l_j}
\end{equation}

When the size of $C$ is 2, the above categorical distribution in Equation \eqref{eq:att-att-iha-l} turns into a Bernoulli distribution:

\begin{equation}\label{eq:att-att-iha-l2}
  l_j = \mathcal{B}\left(T, \alpha_{j1} \right)
\end{equation}

\subsubsection{Location-wise hard attention}\label{sec:att-att-lha}

As mentioned above, for some types of input $X$, like an image, it is not trivial to directly extract items from them. So the location-wise hard attention model is developed which accepts the whole feature map $X$ as input, stochastically picks a sub-region from it, and uses this sub-region to calculate the intermediate code at each decoding step. This kind of location-wise hard attention is analyzed in many previous works \cite{DenilBLF12, larochelle2010learning}, while here we only focus on two recent proposed mechanisms in \cite{MnihHGK14} and \cite{ba-attention-2015}, because the attention models in those works are more general and can be trained end-to-end.

Since the location-wise hard attention model only picks a sub-region of the input $X$ at the decoding step, it makes sense to estimate that the picked region may not correspond to the output item to be predicted. There are two potential solutions for this problem:

\begin{enumerate}
  \item Pick only one input vector at each decoding step.
  \item Make a few \emph{glimpses} for each prediction, which means at each decoding step, pick some sub-regions, process them and update the model one by one, where the subsequent region (glimpse) is acquired based on the previous processed regions (glimpses). And the last prediction is treated as the "real" prediction in this decoding step. The number of glimpses $M$ either can be an arbitrary number (a hyperparameter) or automatically be decided by the model.
\end{enumerate}

With a large enough training dataset, it is reasonable to estimate that method 1 can also find the optimal while the convergence may take longer time because of the limitation mentioned above. In testing, obviously method 1 is faster than method 2, but there may be a sacrifice on the performance. For method 2, it may have better performance in testing, while it also needs longer calculating time. Till now there are not so many works done on how to select an appropriate number of glimpses which leaves it an open problem. However, it is clear that method 1 is just an extreme case of method 2, and in the following sections, we treat the hard attention models as taking $M$ glimpses at each step of prediction.

To keep the notation consistent, here we still use term $l$ as the indicator generated by the attention model to show which sub-region should be extracted, i.e., in the case of location-wise hard attention, the $l_j^m$ indicates the location of the center of the sub-region is about to be picked. In detail, at the $m$-th glimpse in the decoding step $j$, the attention module accepts the input $X$, the previous hidden states of the decoder ($\pmb{h}_j^{m-1}$), and calculates $l_j^m$ by:

\begin{equation}\label{eq:att-att-lha-l}
  l_{j}^m \sim \mathcal{N} \left(f_{att}\left(X, \pmb{h}_j^{m-1} \right) , s \right)
\end{equation}

where $\pmb{h}_j^{0} = \pmb{h}_{j-1} $ and $\mathcal{N} () $ is a normal distribution with a mean $f_{att} \left(X,\pmb{h}_j^{m-1} \right)$ and a standard deviation $s$. And $f_{att}$ usually is a neural network similar to the $f_{att}$ in Equation \eqref{eq:att-att-isa-e}. Then the attention module picks the sub-region centered at location $l_j^m$:

\begin{equation}\label{eq:att-att-lha-Xout}
  X_{out}^m = X_{l_j^m}
\end{equation}

where $X_{l_j^m}$ indicates a sub-region of $X$ centered at $l_j^m$, and $X_{out}^m$ represents the output sub-region at glimpse $m$. Note here that the shape, size of the sub-region, and the standard deviation $s$ in Equation \eqref{eq:att-att-lha-l} are all hyperparameters. One can also let the attention network generate these parameters as the generation of $l_j^m$ in Equation \eqref{eq:att-att-lha-l} \cite{yoo2015attentionnet}. When $X_{out}^m$ is generated, it is fed into the encoder in calculating the code $\pmb{c}^m$ for the decoder, i.e.,

\begin{equation}\label{eq:att-att-lha-cm}
  \pmb{c}^m = \phi_{W_{enc}} \left( X_{out}^m \right)
\end{equation}

and then we have

\begin{equation}\label{eq:att-att-lha-de}
  \begin{bmatrix}
\pmb{\hat{y}}_j^m \\
\pmb{h}_j^m \\ \end{bmatrix} = \phi_{W_{dec}}\left(\pmb{c}^m, \pmb{\hat{y}}_{j}^{m-1}, \pmb{h}_{j}^{m-1} \right)
\end{equation}

After $M$ glimpses:

\begin{equation}\label{eq:att-att-lha-c}
  \pmb{c} = \pmb{c}^M
\end{equation}

\begin{equation}\label{eq:att-att-lha-h}
  \pmb{h}_j = \pmb{h}_j^M
\end{equation}

\begin{equation}\label{eq:att-att-lha-y}
  \pmb{\hat{y}}_j = \pmb{\hat{y}}_j^M
\end{equation}

where $\pmb{h}_j$ is the $j$-th hidden state of the decoder, and $\pmb{\hat{y}}_j$ is the $j$-th prediction.

\subsubsection{Location-wise soft attention}

\begin{figure}[!ht]
  \centering
  \makebox[\textwidth]{\includegraphics[width=\textwidth]{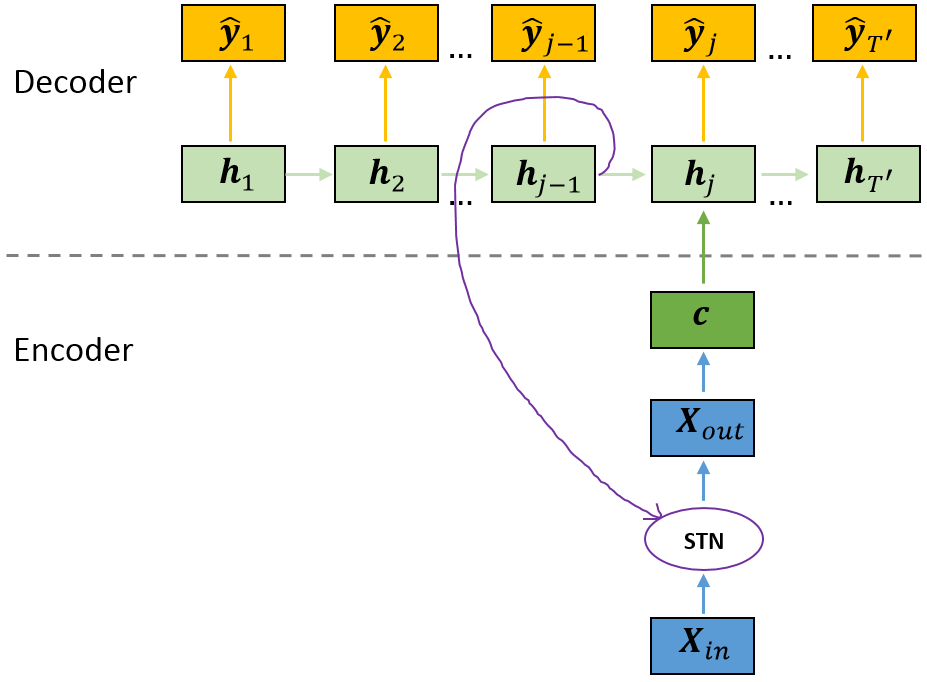}}
  \caption{The location-wise soft attention (STN) based RNN model at decoding step $j$. The purple ellipse represents the STN module which will be described in detail later. $X_{in}$ is the input feature map, and $X_{out}$ is the output feature map.}
  \label{fig:att-att-lsa-ex1}
\end{figure}

As mentioned above, the location-wise soft attention also accepts a feature map $X$ as input, and at each decoding step, a transformed version of the input feature map is generated to calculate the intermediate code. It is firstly proposed in \cite{jaderberg2015spatial} called the spatial transformation network (STN). Instead of the RNN model, the STN in \cite{jaderberg2015spatial} is initially applied on a CNN model, but it can be easily transferred to a RNN model with tiny modifications. To make the explanations more clear, we use $X_{in}$ to represent the input of the STN (where usually $X=X_{in}$), and $X_{out}$ to represent the output of the STN. Figure \ref{fig:att-att-lsa-ex1} gives a visual example of how the STN is embedded into the whole framework, and how it works at decoding step $j$, in which the purple ellipse is the STN module. From Figure \ref{fig:att-att-lsa-ex1}, we can see that after being processed by the STN, the $X_{in}$ is transformed to $X_{out}$, and it is used to calculate the intermediate code $\pmb{c}$ which is then fed to the decoder. The details of how the STN works will be illustrated in the following part of this section.

\begin{figure}[!ht]
  \centering
  \makebox[\textwidth]{\includegraphics[width=\textwidth]{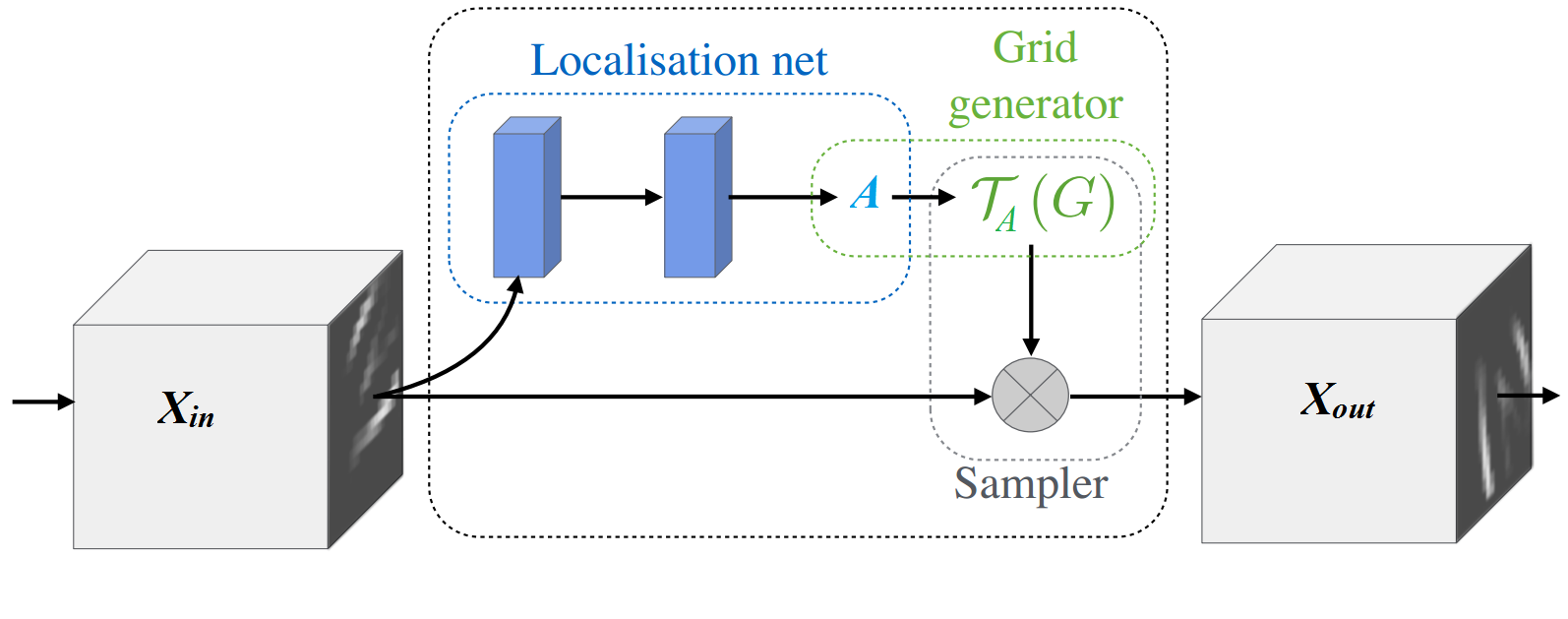}}
  \caption{The structure of STN. The figure is taken from \cite{jaderberg2015spatial}.}
  \label{fig:att-att-lsa-stn}
\end{figure}

Let the shape of $X_{in}$ be $U_{in}\times V_{in}\times Q_{in}$ and the shape of $X_{out}$ be $U_{out}\times V_{out}\times Q_{out}$, where $U$, $V$, $Q$ represent the height, width, and number of channels respectively, so when $X_{in}$ is a color image, its shape should be $U_{in}\times V_{in}\times 3$. The STN works identically on each channel to keep the consistency between the channels, so the number of channels of the input and output are the same ($Q=Q_{in}=Q_{out}$). A typical STN network consists of three parts shown in Figure \ref{fig:att-att-lsa-stn}: localization network, grid generator, and the sampler.

At each decoding time $j$, the localization network $\phi_{W_{loc}}$ takes the input $X_{in}$, the previous hidden state of the decoder $\pmb{h}_{j-1}$ as input, and generates the transformation parameter $A_j$ as output:

\usetagform{fn}
\begin{equation}\label{eq:att-att-lsa-loc}
  A_j=\phi_{W_{loc}}(X_{in}, \pmb{h}_{j-1})
\end{equation}
\footnotetext{Since the original STN in \cite{jaderberg2015spatial} is applied on a CNN, there is no hidden state. As a result, in \cite{jaderberg2015spatial} the location network only takes $X_{in}$ as input, and Equation (\ref{eq:att-att-lsa-loc}) changes to $A_j=\phi_{W_{loc}}(X_{in}) $.}

\usetagform{default}

\begin{figure}[!ht]
  \centering
  \makebox[\textwidth]{\includegraphics[width=\textwidth]{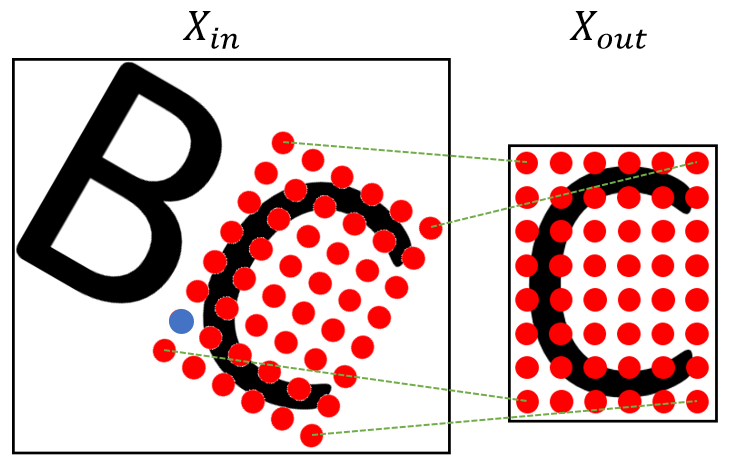}}
  \caption{An example of the transformation $S = \tau_A (G) $, where the left part is $X_{in}$, the right part shows $X_{out}$. The red dots in the output image make up the grid generated by the grid generator as shown in Equation \eqref{eq:att-att-lsa-grid}, and the red (and blue) dots in $X_{in}$ are the elements of $S$ as shown in Equation \eqref{eq:att-att-lsa-S}. The blue dot in $X_{in}$ will be used to illustrate how the sampling works later.}
  \label{fig:att-att-lsa-transgrid}
\end{figure}

Then the transformation $\tau$ parameterized by $A_j$ is applied on a mesh grid $G$ generated by the grid generator to produce a feature map $S$ which indicates how to select pixels\footnotemark{} from $X_{in}$ and map them to $X_{out}$. In detail, the grid $G$ consists of pixels of the output feature map $X_{out}$, i.e.,

\footnotetext{Here the pixel means the element of the feature map $X_{in}$, which does not have to be an image.}

\begin{equation}\label{eq:att-att-lsa-grid}
  G= \left\{ G_i \right\} = \left\{ \left(x^{X_{out}}_{1,1}, y^{X_{out}}_{1,1}\right) , \left(x^{X_{out}}_{1,2}, y^{X_{out}}_{1,2}\right), \dots, \left(x^{X_{out}}_{U_{out},V_{out}}, y^{X_{out}}_{U_{out},V_{out}}\right) \right\}
\end{equation}

where $x$ and $y$ represent the coordinates of the pixel, and $\left(x_{1,1}^{\left(X_{out} \right)},y_{1,1}^{\left(X_{out} \right)} \right)$ is the pixel of $X_{out}$ which locates at coordinate $(1,1)$. And we have

\begin{equation}\label{eq:att-att-lsa-Si}
  S_i = \tau_{A_j} (G_i)
\end{equation}

\begin{equation}\label{eq:att-att-lsa-S}
  S = \left\{ S_i \right\} = \left\{ \left(x^{S}_{1,1}, y^{S}_{1,1}\right), \left(x^{S}_{1,2}, y^{S}_{1,2}\right), \dots, \left(x^{S}_{U_{out},V_{out}}, y^{S}_{U{out},V_{out}}\right) \right\}
\end{equation}

where each $S_i$ shows the coordinates in the input feature map that defines a sampling position. Figure \ref{fig:att-att-lsa-transgrid} gives a visual example illustrating how the transformation and the grid generator work. In Figure \ref{fig:att-att-lsa-transgrid}, the grid $G$ consists of the red dots in the right part of the figure. Then a transformation (the dashed green lines in the figure) is applied on $G$ to generate $S$, which is the red (and blue) dots in the left part of the figure. Then $S$ indicates the positions to perform the sampling.

Actually, $\tau$ can be any transformations, for example, the affine transformation, the plane projective transformation, or the thin plate spline. In order to make the descriptions above more clear, we assume $\tau$ is a 2D affine transformation, so $A_j$ is a matrix consists of 6 elements:

\begin{equation}\label{eq:att-att-lsa-affine}
  A_j = \begin{bmatrix}
    a_{1,1} & a_{1,2} & a_{1,3}\\
    a_{2,1} & a_{2,2} & a_{2,3}
\end{bmatrix}
\end{equation}

Then Equation \eqref{eq:att-att-lsa-Si} can be rewritten as:

\begin{equation}\label{eq:att-att-lsa-affinetrans}
  S_i = \left( \begin{array}{c}
    x^{S}_{i} \\
    y^{S}_{i}
\end{array} \right) = \tau_{A_j} \left( G_i \right) = A_j \left( \begin{array}{c}
    x^{X_{out}}_{i} \\
    y^{X_{out}}_{i} \\
    1
\end{array} \right) = \begin{bmatrix}
    a_{1,1} & a_{1,2} & a_{1,3}\\
    a_{2,1} & a_{2,2} & a_{2,3}
\end{bmatrix}  \left( \begin{array}{c}
    x^{X_{out}}_{i} \\
    y^{X_{out}}_{i} \\
    1
\end{array} \right)
\end{equation}

\begin{figure}[!ht]
  \centering
  \makebox[\textwidth]{\includegraphics{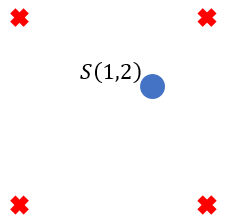}}
  \caption{A visual example of the bilinear interpolation. The blue dot is an element in $S$ which indicates the sampling position. In detail, it corresponds to the blue dot in Figure \ref{fig:att-att-lsa-transgrid}, where $S(1,2)$ means its coordinate is $(1,2)$. The four red intersection symbols represent the pixels in $X_{in}$ which surround the sampling position.}
  \label{fig:att-att-lsa-bilinear}
\end{figure}

The last step is the sampling which is operated by the sampler. Because $S$ is calculated by a transformation, $S_i$ does not always correspond exactly to the pixels in $X_{in}$, hence a sampling kernel is applied to map the pixels in $X_{in}$ to $X_{out}$:

\begin{equation}\label{eq:att-att-lsa-sampling}
  X^q_{out,i} = \sum^{U_{in}}_{u} \sum^{V_{in}}_{v} X^q_{in,u,v} k \left( x^{S}_i - v \right) k \left( y^{S}_i - u \right) \quad \forall i \in [1, 2, \dots, U_{out}V_{out}] \quad \forall q \in [1,2,\dots,Q]
\end{equation}

where $k$ can be any kernel as long as it is partial differentiable with respect to both $x_i^S$ and $y_i^S$, for example, one of the most widely used sampling kernel for images is the bilinear interpolation:

\begin{equation}\label{eq:att-att-lsa-bilinear}
  \begin{split}
X^q_{out,i} = \sum^{U_{in}}_{u} \sum^{V_{in}}_{v} X^q_{in,u,v} \max \left(0, 1 - \left| x^{S}_i - v \right| \right) \max \left(0, 1 - \left| y^{S}_i - u \right| \right) \\
\quad \forall i \in [1, 2, \dots, U_{out}V_{out}] \quad \forall q \in [1, 2, \dots, Q]
\end{split}
\end{equation}

when the coordinates of $X_{in}$ and $X_{out}$ are normalized, i.e., $\left(x_{1,1}^{X_{in}},y_{1,1}^{X_{in} } \right)=\left(-1,-1\right)$ and $\left(x_{U_{in},V_{in}}^{X_{in} },y_{U_{in},V_{in}}^{X_{in}} \right)=\left(+1,+1\right)$. A visual example of the bilinear interpolation is shown in Figure \ref{fig:att-att-lsa-bilinear}, where the blue dot represents a sampling position, and its value is calculated by a weighted sum of its surrounding pixels as shown in Equation \eqref{eq:att-att-lsa-bilinear}. In Figure \ref{fig:att-att-lsa-bilinear}, the surrounding pixels of the sampling position are represented by four red intersection symbols.

Now if we review the whole process described above, the STN:

\begin{enumerate}
  \item Accepts a feature map, and the previous hidden state of the decoder as input.
  \item Generates parameters for a pre-defined transformation.
  \item Generates the sampling grid and calculates the corresponding sampling positions in the input feature map based on the transformation.
  \item Calculates the output pixel values by a pre-defined sampling kernel.
\end{enumerate}

Therefore, for an input feature map $X_{in}$, an output feature map $X_{out}$ is generated which has the ability to only focus on some interesting parts in $X_{in}$. For instance, the affine transformation defined in Equation \eqref{eq:att-att-lsa-affinetrans} allows translation, rotation, scale, skew, and cropping, which is enough for most of the image related tasks. The whole process introduced above, i.e., feature map transformation, grid generation, and sampling, is inspired by the standard texture mapping in computer graphics.

At last the $X_{out}$ is fed into the encoder to generate the intermediate code $\pmb{c}$:

\begin{equation}\label{eq:att-att-lsa-c}
  \pmb{c} = \phi_{W_{enc}} (X_{out})
\end{equation}

\subsection{Learning of the attention based RNN models}

As mentioned in the introduction, the difference between the soft and hard attention mechanisms in optimization is that the soft attention module is differentiable with respect to its parameters so the standard gradient ascent/decent can be used for optimization. However, for the hard attention, the model is non-differentiable and the techniques from the reinforcement learning are applied for optimization.

\subsubsection{Soft attention}

For the item-wise soft attention mechanism, when $f_{att}$ in Equation \eqref{eq:att-att-isa-e} is differentiable with respect to its inputs, it is clear that the whole attention model is differentiable with respect to $W_{att}$, as well as the whole RNN model is differentiable with respect to $\theta=\left[W_{enc},W_{dec},W_{att} \right]$. As a result, the same sum of log-likelihood in Equation \eqref{eq:att-sum-ll} is used as the objective function for the learning, and the gradient ascent is used to maximize the objective.

The location-wise soft attention module (STN) is also differentiable with respect to its parameters if the location network $\phi_{W_{loc} }$, the transformation $\tau$ and the sampling kernel $k$ are carefully selected to ensure their gradients with respect to their inputs can be defined. For example, when using a differentiable location network $\phi_{W_{loc} }$, the affine transformation (Equation \eqref{eq:att-att-lsa-affine}, Equation \eqref{eq:att-att-lsa-affinetrans}), and the bilinear interpolation (Equation \eqref{eq:att-att-lsa-bilinear}) as the sampling kernel, we have:

\begin{equation}\label{eq:att-att-learning-dxoutdxin}
  \frac {\partial X^q_{out, i}} { \partial X^q_{in,u,v}} = \sum^{U_{in}}_{u} \sum^{V_{in}}_{v} \max \left(0, 1 - \left| x^{S}_i - v \right| \right) \max \left(0, 1 - \left| y^{S}_i - u \right| \right)
\end{equation}

\begin{equation}\label{eq:att-att-learning-dxoutdxs}
  \frac {\partial X^q_{out, i}} { \partial x^S_i} = \sum^{U_{in}}_{u} \sum^{V_{in}}_{v} X^q_{in, u, v} \max \left(0, 1 - \left| y^{S}_i - u \right| \right) \left\{ \begin{split}
  0 &\quad \mbox{ if } \left| v-x^S_i \right| \ge 1 \\
  1 &\quad \mbox{ if } v \ge x^S_i  \\
  -1 &\quad \mbox{ if } v < x^S_i
\end{split}
\right.
\end{equation}

\begin{equation}\label{eq:att-att-learning-dxoutdxy}
  \frac {\partial X^q_{out, i}} { \partial y^S_i} = \sum^{U_{in}}_{u} \sum^{V_{in}}_{v} X^q_{in, u, v} \max \left(0, 1 - \left| x^{S}_i - v \right| \right) \left\{ \begin{split}
  0 &\quad \mbox{ if } \left| u-y^S_i \right| \ge 1 \\
  1 &\quad \mbox{ if } u \ge y^S_i  \\
  -1 &\quad \mbox{ if } u < y^S_i
\end{split}
\right.
\end{equation}

\begin{equation}\label{eq:att-att-learning-dxsda11}
  \frac {\partial x^S_i} {\partial a_{1,1}} = x^{X_{out}}_i
\end{equation}

\begin{equation}\label{eq:att-att-learning-dxsda12}
  \frac {\partial x^S_i} {\partial a_{1,2}} = y^{X_{out}}_i
\end{equation}

\begin{equation}\label{eq:att-att-learning-dxsda13}
  \frac {\partial x^S_i} {\partial a_{1,3}} = 1
\end{equation}

\begin{equation}\label{eq:att-att-learning-dysda21}
  \frac {\partial y^S_i} {\partial a_{2,1}} = x^{X_{out}}_i
\end{equation}

\begin{equation}\label{eq:att-att-learning-dysda22}
  \frac {\partial y^S_i} {\partial a_{2,2}} = y^{X_{out}}_i
\end{equation}

\begin{equation}\label{eq:att-att-learning-dysda23}
  \frac {\partial y^S_i} {\partial a_{2,3}} = 1
\end{equation}

Now the gradients of $W_{loc}$ ($\frac{\partial A_j}{\partial W_{loc}}$) can be easily derived from Equation (\ref{eq:att-att-lsa-loc}). As a result, the entire STN is differentiable with respect to its parameters. As in the item-wise soft attention, the objective function is the sum of the log-likelihood, and the gradient ascent is used for optimization.

\subsubsection{Hard attention}

As shown in Section \ref{sec:att-att-iha} and Section \ref{sec:att-att-lha}, the hard attention mechanism stochastically picks an item or a sub-region from the input. In this case, the gradients with respect to the picked item/region are zero because they are discrete. The zero gradients cannot be used to maximize the sum of the log-likelihood by the standard gradient ascent. In general, this is a problem of training a neural network with discrete values/units, and here we just introduced the method applied in \cite{MnihHGK14, ba-attention-2015, XuBKCCSZB15} which employs the techniques from the reinforcement learning. The learning methods for the item-wise and location-wise hard attention are essentially the same, but the item-wise hard attention implicitly sets the number of glimpse as 1. In this section, we just make the number of glimpses as $M$.

Instead of the raw log-likelihood in Equation \eqref{eq:att-ll}, a new objective $L'_j$ is defined which is a variational lower bound on the log-likelihood $\log p\left(\pmb{y}_i|X, \pmb{y}_1, \pmb{y}_2, \dots, \pmb{y}_{i-1}, \theta   \right)$ \footnotemark{} in Equation \eqref{eq:att-ll} parameterized by $\theta$ ($\theta=\left[ W_{enc}, W_{dec}, W_{att} \right]$). And we have

\footnotetext{For notational simplicity, we ignore the $\pmb{y}_1, \pmb{y}_2, \dots, \pmb{y}_{i-1}, \theta$ in the log-likelihood later, so $\log p\left(\pmb{y}_i|X, \pmb{y}_1, \pmb{y}_2, \dots, \pmb{y}_{i-1}, \theta   \right)=\log p\left(\pmb{y}_i|X\right)$}

\begin{equation}\label{eq:att-att-learning-ha-lj}
  \begin{aligned}
\mbox{log-likelihood} = L_j &= \log p(\pmb{y}_j|X) \\
&\geqslant \log \sum_{l_j} p(l_j|X)p(\pmb{y}_j|l_j,X) \\
&=\sum_{l_j} p(l_j|X)\log p(\pmb{y}_j|l_j,X) \\
&=\sum_{l_j} p(l_j|\pmb{x}^l)\log p(\pmb{y}_j|l_j,\pmb{x}^l) = L'_j
\end{aligned}
\end{equation}

Then the derivative of $L'_j$ is:

\begin{equation}\label{eq:att-att-learning-ha-dlj}
  \begin{aligned}
\frac {\partial L'_j} {\partial \theta} &= \sum_{l_j} \left( p(l_j|\pmb{x}^l) \frac {\partial \log p(\pmb{y}_j|l_j,\pmb{x}^l)} {\partial \theta} + \log p(\pmb{y}_j|l_j,\pmb{x}^l) \frac {\partial p(l_j|\pmb{x}^l)} {\partial \theta} \right) \\
&= \sum_{l_j} p(l_j|\pmb{x}^l) \left( \frac {\partial \log p(\pmb{y}_j|l_j,\pmb{x}^l)} {\partial \theta} + \log p(\pmb{y}_j|l_j,\pmb{x}^l) \frac {\partial \log p(l_j|\pmb{x}^l)} {\partial \theta} \right)
\end{aligned}
\end{equation}

As shown above, $l_j$ is generated from a distribution (Equation \eqref{eq:att-att-iha-l}, Equation \eqref{eq:att-att-iha-l2}, or Equation \eqref{eq:att-att-lha-l}), which indicates that $p(l_j |x^l )$ and $\frac{\partial \log p\left( l_j | x^l \right)}{\partial \theta}$ can be estimated by a Monte Carlo sampling as demonstrated in \cite{Williams92}:

\begin{equation}\label{eq:att-att-learning-ha-dljMC}
  \frac {\partial L'_{j}} {\partial \theta} \approx \frac {1} {M} \sum_{m=1}^{M} \left( \frac {\partial \log p(\pmb{y}_j|l_j^m,\pmb{x}^l)} {\partial \theta} + \log p(\pmb{y}_j|l_j^m,\pmb{x}^l) \frac {\partial \log p(l_j^m|\pmb{x}^l)} {\partial \theta} \right)
\end{equation}

The whole learning process described above for the hard attention is equivalent to the REINFORCE learning rule in \cite{Williams92}, and from the reinforcement learning perspective, after each step, the model can get a reward from the environment. In Equation \eqref{eq:att-att-learning-ha-dljMC}, the $\log p(\pmb{y}_j|l_j^m,\pmb{x}^l)$ is used as the reward $R_j$. But in the reinforcement learning, the reward can be assigned to an arbitrary value. Depending on the tasks to be solved, here are some widely used schemes:

\begin{itemize}
  \item Set the reward to be exactly $R_j = \log p(\pmb{y}_j|l_j^m,\pmb{x}^l)$.
  \item Set the reward to be a real value proportional to $\log p(\pmb{y}_j|l_j^m,\pmb{x}^l)$, which means $R_j = \beta \log p(\pmb{y}_j|l_j^m,\pmb{x}^l)$, and $\beta$ is a hyperparameter.
  \item Set the reward as a zero/one discrete value:
  \begin{equation}\label{eq:att-att-learning-ha-r3}
    R_j = \left\{ \begin{array}{ll}
         1 & \pmb{y}_j = \underset{\pmb{y}_j} {\mathrm{argmax}} \log p(\pmb{y}_j|l_j^m, \pmb{x}^l)\\
         0 & \mbox{otherwise}.\end{array} \right.
  \end{equation}
\end{itemize}

Then Equation \eqref{eq:att-att-learning-ha-dljMC} can be written as:

\begin{equation}\label{eq:att-att-learning-ha-dljMC-r}
  \frac {\partial L'_j} {\partial \theta} \approx \frac {1} {M} \sum_{m=1}^{M} \left( \frac {\partial \log p(\pmb{y}_j|l_j^m,\pmb{x}^l)} {\partial \theta} + R_j \frac {\partial \log p(l_j^m|\pmb{x}^l)} {\partial \theta} \right)
\end{equation}

Now as shown in \cite{MnihHGK14}, although Equation \eqref{eq:att-att-learning-ha-dljMC-r} is an unbiased estimation of the real gradient of Equation \eqref{eq:att-att-learning-ha-dlj}, it may have high variance because of the unbounded $R_j$. As a result, usually a variance reduction item $b$ is added to the equation:

\begin{equation}\label{eq:att-att-learning-ha-dljMC-rb}
  \frac {\partial L'_j} {\partial \theta} \approx \frac {1} {M} \sum_{m=1}^{M} \left( \frac {\partial \log p(\pmb{y}_j|l_j^m,\pmb{x}^l)} {\partial \theta} + (R_j - b_j) \frac {\partial \log p(l_j^m|\pmb{x}^l)} {\partial \theta} \right)
\end{equation}

where $b$ can be calculated in different ways which are discussed in \cite{ba-attention-2015, BaGSF15, MnihHGK14, WeaverT01, XuBKCCSZB15}. A common approach is to set $ b_j=\mathop{\mathbb{E}}(R) $. By using the variance reduction variable $b$, the training becomes faster and more robust. At last, a hyperparameter $\lambda$ can be added to balance the two gradient components:

\begin{equation}\label{eq:att-att-learning-ha-dljMC-rba}
  \frac {\partial L'_j} {\partial \theta} \approx \frac {1} {M} \sum_{m=1}^{M} \left( \frac {\partial \log p(\pmb{y}_j|l_j^m,\pmb{x}^l)} {\partial \theta} + \lambda(R_j - b_j) \frac {\partial \log p(l_j^m|\pmb{x}^l)} {\partial \theta} \right)
\end{equation}

With the gradients in Equation \eqref{eq:att-att-learning-ha-dljMC-rba}, now the hard attention based RNN model can also be trained by gradient ascent. For the whole sequence $Y$, we have the new objective function $L'(Y,X,θ)$:

\begin{equation}\label{eq:att-att-learning-ha-obj}
  \begin{aligned}
L'(Y,X, \theta) &= \sum_{j=1}^{T'} \sum_{l_{j}} p \left(l_{j}|\pmb{x}^l \right) \log p \left(\pmb{y}_j|l_{j}, \pmb{x}^l \right) \\
&= \sum_{j=1}^{T'} \sum_{l_{j}} p \left(l_{j}|X \right) \log p \left(\pmb{y}_j|l_{j}, X \right) \\
&\leqslant \sum_{j=1}^{T'} \log \sum_{l_{j}} p(\pmb{y}_j|X) \\
& = \sum_{j=1}^{T'} \log p(\pmb{y}_j|X) \\
& = \sum_{j=1}^{T'} \log p\left(\pmb{y}_j|X, \pmb{y}_1, \pmb{y}_2, \dots, \pmb{y}_{j-1}, \theta \right) = \mbox{sum of the log-likelihood}
\end{aligned}
\end{equation}

\subsection{Summary}

In this section, we firstly give a brief introduction of the neural network, the CNN, and the RNN. Then we discuss the common RNN model for sequence to sequence problem. And the detailed descriptions of four types of attention mechanisms are given: item-wise soft attention, item-wise hard attention, location-wise hard attention, and location-wise soft attention, followed by how to optimize the attention based RNN models.

%% file: application.tex
\section{Applications of the attention based RNN model in computer vision} \label{sec:app}

In this section, we firstly give a brief overview of the differences when one uses different attention based RNN models. And then two applications in computer vision which apply the attention based RNN model are introduced.

\subsection{Overview}

\begin{itemize}
  \item As mentioned above, the item-wise attention model requires that the input sequence consists of explicit items, which is not trivial for some types of input, like an image input. A simple solution is to manually extract some patches from the input images and treat them as a sequence of item. For the extraction method, one can either extract some fixed size patches from some pre-defined locations, or apply other object proposal methods, like \cite{UijlingsSGS13}.
  \item The location-wise attention model can directly accept a feature map as input which avoids the "items extraction" step mentioned above. And if compared to human perception, the location-wise attention is more appealing. Because when human being sees an image, he will not manually divide the image into some patches and calculate the weights for them before recognizing the objects in the image. Instead, he will directly focus on an object and its surroundings. That is exactly what the location-wise attention does.
  \item The item-wise attention model has to calculate the codes for all items in the input sequence, which is less efficient than the location-wise attention model.
  \item The soft attention is differentiable with respect to its inputs, so the model can be trained by optimizing the sum of the log-likelihood using gradient ascent. However, the hard attention is non-differentiable, usually the techniques from the reinforcement learning are applied to maximize a variational lower bound of the log-likelihood, and it makes sense to estimate there will be a sacrifice on the performance.
  \item The STN keeps both the advantages of the location-wise attention and the soft attention, where it can directly accept a feature map as input and optimize the raw sum of log-likelihood. However, it is just recently proposed, and originally applied on CNN. Currently, there are not enough applications of the STN based RNN model.
\end{itemize}

\subsection{Image classification and object detection}

Image classification is one of the most classical and fundamental applications in computer vision area, where each image has a (some) label(s), and the model needs to learn how to predict the label given a new input image. Now the most popular and successful model for image classification task is the CNN. While as mentioned above, the CNN takes one fixed size vector as input, which has some disadvantages:

\begin{itemize}
  \item When the input image is larger than the input size accepted by the CNN, either the image needs to be rescaled to a smaller size to meet the requirements of the model, or the image needs to be cut into some subparts to be processed one by one. If the image is rescaled smaller, there are some sacrifices on the details of the image, which may harm the classification performance. On the other hand, if the image is cut into some patches, the amount of computation will scale linearly with the size of the image.
  \item The CNN has a degree of spatial transformation invariance build-in, while when the image is noisy or the object indicated by the label only occupies a small region of the input image, a CNN model may not still keep satisfied performance.
\end{itemize}

As a result, \cite{MnihHGK14} and \cite{ba-attention-2015} propose the RNN based image classification models with the location-wise hard attention mechanism. As explained above, the location-wise hard attention based model stochastically picks a sub-region from the input image to generate the intermediate code. So the models in \cite{MnihHGK14} and \cite{ba-attention-2015} can not only make prediction of the image label, but also localize the position of the object. This means the attention based RNN model integrates the image classification and object detection into a single end-to-end trainable model, which is another advantage compared to the CNN based object detection model. If one wants to apply a convolution network to solve the object detection problem, he has to use a separate model to propose the potential locations of the objects, like \cite{UijlingsSGS13}, which is expensive.

\begin{figure}[!ht]
  \centering
  \makebox[\textwidth]{\includegraphics[width=0.7\textwidth]{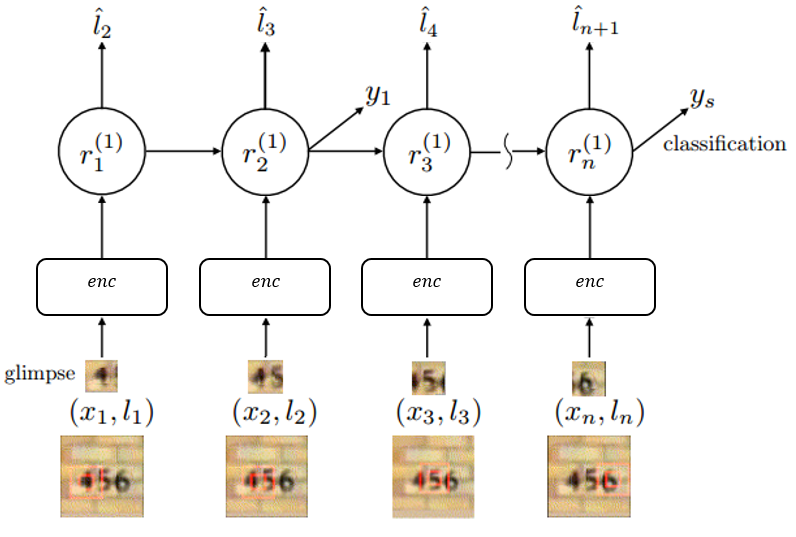}}
  \caption{The model proposed in \cite{MnihHGK14}. $enc$ is the encoder, $r^{(1)}$ is the decoder and the attention model. $\hat{l}$ is equivalent to the output of $f_{att}()$ in Equation \eqref{eq:att-att-lha-l}, which is used to generate $l$ as shown in Equation \eqref{eq:att-att-lha-l}. $x$ is the sub-region of the whole image extracted based on $l$, and $y_s$ is the predicted label for the image. The figure is taken from \cite{ba-attention-2015}, and some modifications are made to fit the model in \cite{MnihHGK14}.}
  \label{fig:app-cls-MnihHGK14}
\end{figure}

In this section, we will briefly introduce and analyze the models proposed in \cite{MnihHGK14}, \cite{ba-attention-2015}, and their extensions. A brief structure of the RNN model in \cite{MnihHGK14} is shown in Figure \ref{fig:app-cls-MnihHGK14}. In this model, $enc$ is the encoder neural network taking a patch $x$ of the input image to generate the intermediate code, where $x$ is cut from the raw image based on a location value $l$ generated by the attention network. In detail, $l$ is generated by Equation \eqref{eq:att-att-lha-l}, and $x$ is extracted by Equation \eqref{eq:att-att-lha-Xout}. Then the decoder accepts the intermediate code as input to make the potential prediction. Here in Figure \ref{fig:app-cls-MnihHGK14} the decoder and the attention network are put in the same network $r^{(1)}$.

The experiments in \cite{MnihHGK14} compare the performances of the proposed model with some non-recurrent neural networks, and the results show the superiority of the attention based RNN model to the non-recurrent neural networks with similar number of parameters, especially on the datasets with noise. For the details of the experiments, one can refer to Section 4.1 in \cite{MnihHGK14}.

\begin{figure}[!ht]
  \centering
  \makebox[\textwidth]{\includegraphics[width=0.7\textwidth]{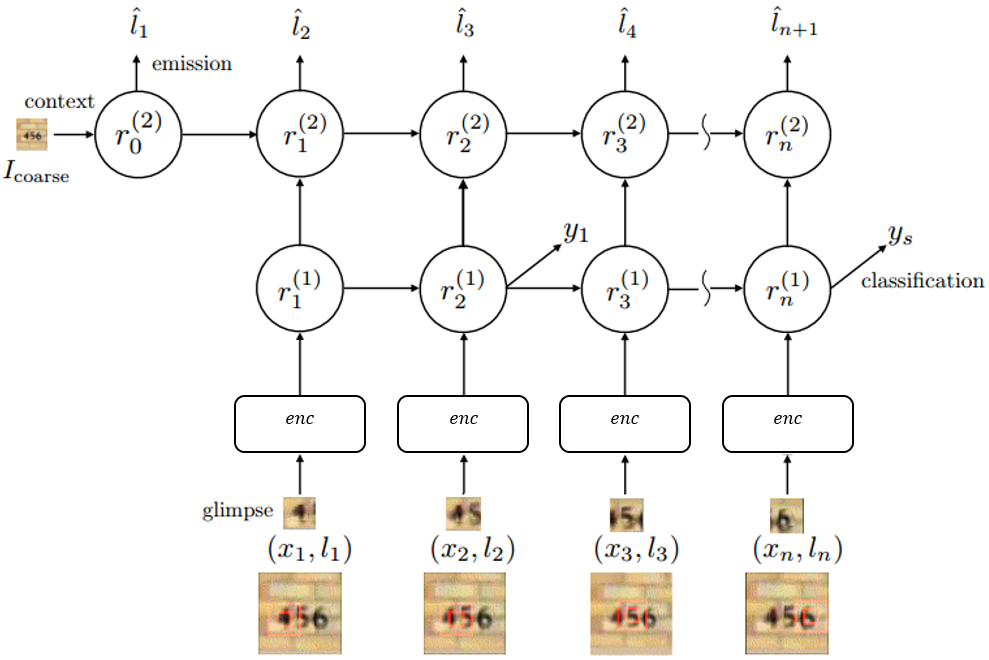}}
  \caption{The model proposed in \cite{ba-attention-2015}. $enc$ is the encoder, $r^{(1)}$ is the decoder network, and $r^{(2)}$ is the attention network. $I_{coarse}$ is a context vector extracted from the whole input image to generate the location of the first glimpse. The other terms are the same as in Figure \ref{fig:app-cls-MnihHGK14}. The figure is taken from \cite{ba-attention-2015}, and some modifications are made.}
  \label{fig:app-cls-ba-attention-2015}
\end{figure}

However, the original model shown in Figure \ref{fig:app-cls-MnihHGK14} is very simple, where $enc$ is a two-layer neural network, and $r^{(1)}$ is a three-layer neural network. Besides, the first glimpse ($l_1$ in Figure \ref{fig:app-cls-MnihHGK14}) is assigned manually. And all the experiments are only conducted on some toy datasets. There is no evidence to prove the generalization power of the model in Figure \ref{fig:app-cls-MnihHGK14} to some real-world problems. \cite{ba-attention-2015} proposes an extended version of model as shown in Figure \ref{fig:app-cls-ba-attention-2015} by firstly making the network deeper, and secondly using a context vector to obtain a better first glimpse.

The encoder ($enc$) in \cite{ba-attention-2015} is deeper, i.e., it consists of three convolutional layers and one fully connected layer. Another big difference between the model in Figure \ref{fig:app-cls-ba-attention-2015} and the model in Figure \ref{fig:app-cls-MnihHGK14} is that the model in Figure \ref{fig:app-cls-ba-attention-2015} adds an independent layer $r^{(2)}$ as the attention network in order to make the first glimpse as accurate as possible, i.e., the model extracts a context vector from the whole image ($I_{coarse}$ in the figure), and feeds it to the attention model to generate the first potential location $l_1$. The same context vector of the whole image is not fed into the decoder network $r^{(1)}$, because \cite{ba-attention-2015} observes that if so, the predicted label are highly influenced by the information of the whole image. Besides, both $r^{(1)}$ and $r^{(2)}$ are recurrent, while in Figure \ref{fig:app-cls-MnihHGK14}, only one hidden state exists.

\begin{figure}[!ht]

\begin{subfigure}[t]{0.5\textwidth}
    \includegraphics[width=0.95\linewidth]{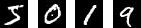}
    \caption{Four examples taken from the MNIST dataset.}
    \label{fig:app-cls-MNIST-ex}
\end{subfigure}
\begin{subfigure}[t]{0.5\textwidth}
    \includegraphics[width=0.95\linewidth]{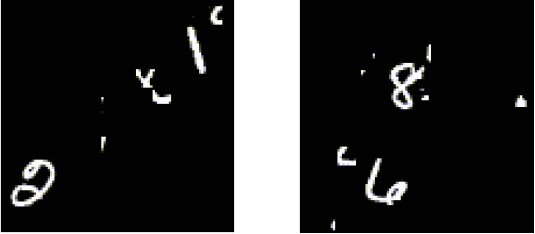}
    \caption{Two examples taken from the MNIST pairs dataset \cite{MnihHGK14}.}
    \label{fig:app-cls-MNISTpairs-ex}
\end{subfigure}

\caption{Examples of MNIST and MNIST pairs datasets.}
\label{fig:app-cls-MNISTpairs}
\end{figure}

Both two models introduced above are evaluated on a dataset called "MNIST pairs" which is generated from the MNIST dataset. MNIST \cite{MNIST} is a handwritten digits dataset, which contains 70000 28\texttimes28 binary images (Figure \ref{fig:app-cls-MNIST-ex}). The MNIST pairs dataset randomly puts two digits into a 100\texttimes100 image with some additional noise in the background \cite{MnihHGK14} (Figure \ref{fig:app-cls-MNISTpairs-ex}). The results are shown in Table \ref{table:app-imcls-MNIST}. It is clear that the performance of \cite{ba-attention-2015} is better, and the context vector indeed makes some improvements by suggesting a more accurate first glimpse.

\begin{table}[]
\centering
\caption{Error rates on the MNIST pairs image classification dataset.}
\label{table:app-imcls-MNIST}
\begin{tabular}{|l|l|}
\hline
Model & Test error   \\ \hhline{|==|}
\cite{MnihHGK14}      & 9\%          \\ \hline
\cite{ba-attention-2015} without the context vector      & 7\%          \\ \hline
\cite{ba-attention-2015}     & \textbf{5\%} \\ \hline
\end{tabular}
\end{table}

The model in \cite{ba-attention-2015} are also tested on multi-digit street view house number (SVHN) \cite{netzer2011reading} sequence recognition task, and the results show that with the same level of error rates, the number of parameters to be learned, and the training time of the attention based RNN model are much less than the state-of-the-art CNN methods.

However nowadays both MNIST and SVHN are toy datasets with constraints environments, i.e., both of them are about digits classification and detection. Besides, their sizes are relatively small compared to other popular image classification/detection datasets, like the ImageNet. \cite{SermanetFR14} further extends the model in \cite{ba-attention-2015} with only a few modifications and applies it to a real-world image classification task: Stanford Dogs fine-grained categorization task \cite{khosla2011novel}, in which the images have larger clutter, occlusion, and variations in pose. The biggest change made in \cite{SermanetFR14} is that a part of the GoogLeNet \cite{SzegedyLJSRAEVR15} is used as the encoder, which is a pre-trained CNN on ImageNet dataset. The performance shown in Table \ref{table:app-imcls-dog} indicates that the attention based RNN model indeed can be applied to real-world datasets while still keep a competitive performance.

\begin{table}[]
\centering
\caption{The classification performance. * represents the model uses additional bounding boxes around the dogs in training and testing.}
\label{table:app-imcls-dog}
\begin{tabular}{|l|l|}
\hline
Model & \begin{tabular}[c]{@{}l@{}}Mean\\ accuracy (MA) \cite{ChaiLZ13}\end{tabular} \\ \hhline{|==|}
\cite{YangBWS12}*      & 0.38                                                                 \\ \hline
\cite{ChaiLZ13}*      & 0.46                                                                 \\ \hline
\cite{gavves2013fine}*      & 0.50                                                                 \\ \hline
GoogLeNet 96\texttimes96 (the encoder in \cite{SermanetFR14})      & 0.42                     \\ \hline
RNN with location-wise hard attention \cite{SermanetFR14}      & \textbf{0.68}      \\ \hline
\end{tabular}
\end{table}

\subsection{Image caption}

Image caption is a very challenging problem: by given an input image, the system needs to generate a natural language sentence to describe the content of the image. The classical way of solving this problem is by dividing the problem into some sub-problems, like object detection, objects-words alignment, sentence generation by the template, etc., and solving them independently. This process will obviously make some sacrifices on the performance. With the development and success of the recurrent network in machine translation area, the image caption problem can also be treated as a machine translation problem, i.e., the image can also be seen as a language, and the system just translate it to another language (natural language, like English). The RNN based image caption system is recently proposed in \cite{VinyalsTBE15} (Neural Image Caption, NIC) which perfectly fits our encoder-decoder RNN framework without the attention mechanism.

\begin{figure}[!ht]
  \centering
  \makebox[\textwidth]{\includegraphics[width=\textwidth]{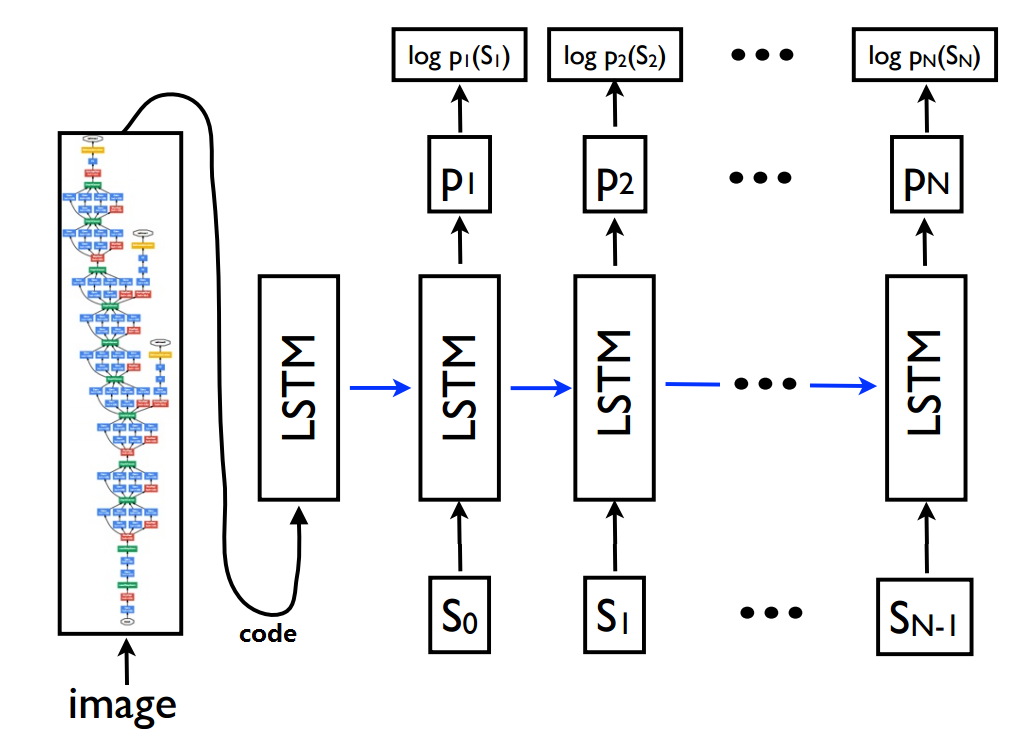}}
  \caption{NIC. The image is firstly processed by a pre-trained CNN and a code is generated, and then the code is decoded by a LSTM unit. $S_i,i=1,2, \dots ,N$ are the word in the image caption sentence. $P_i,i=1,2, \dots ,N$ are the log-likelihoods of the predicted words and the ground truth words. The figure is taken from \cite{VinyalsTBE15}.}
  \label{fig:app-cap-NIC}
\end{figure}

The general structure of the NIC is shown in Figure \ref{fig:app-cap-NIC}. With an input image, the same trick mentioned above is applied: a pre-trained convolutional network is used as the encoder, and the output of a fully-connected layer is the intermediate code. Then it is followed by a recurrent network serving as the decoder to generate the natural language caption, and in NIC, the LSTM unit is used. Compared to the pervious methods which decompose the image caption problem into some sub-problems, the NIC is end-to-end, and is directly trained on the raw data, so the whole system is much simpler and can keep all information of the input data. The experimental results show the superiority of the NIC compared to the traditional methods.

However, one problem for the NIC is that only a single image representation (the intermediate code) is obtained by the encoder from the whole input image, which is counterintuitive, i.e., when human beings describe an image by natural language, usually some objects or some salient areas are focused on one by one. So it is natural to import the ability of "focusing" by applying the attention based RNN model.

\cite{XuBKCCSZB15} adds the item-wise attention mechanisms into the NIC system. The encoder is still a pre-trained CNN, while instead of the fully-connected layer, the output of a convolutional layer is used to calculate the code set $C$ in Equation \eqref{eq:att-att-isa-C}. In detail, at each decoding step the whole image is fed into the pre-trained CNN, and the output of the CNN is a 2D feature map. Then some pre-defined 14\texttimes14 sub-regions of the feature map are extracted as the items in $C$. \cite{XuBKCCSZB15} implements both the item-wise soft and the item-wise hard attention mechanisms. For the item-wise soft attention, at each decoding step a weight for each code in $C$ is calculated by Equation \eqref{eq:att-att-isa-e} and Equation \eqref{eq:att-att-isa-alpha}, and then an expectation of all codes in $C$ (Equation \eqref{eq:att-att-isa-finalc}) is used as the intermediate code. For the item-wise hard attention, the same weights for all codes in $C$ are calculated and the index of the code to be picked is generated by Equation \eqref{eq:att-att-iha-l}.

By using the attention mechanism, \cite{XuBKCCSZB15} reports improvements compared to the common RNN models without attention mechanism including NIC. However, both NIC and \cite{XuBKCCSZB15} have participated in the MS COCO Captioning Challenge 2015\footnote{\label{foot:app-cap-coco}MS COCO Captioning Challenge 2015, http://mscoco.org/dataset/\#captions-challenge2015}, and the competition gives opposite results. MS COCO Captioning Challenge 2015 is an image caption competition running on one of the biggest image caption datasets: Microsoft COCO \cite{LinMBHPRDZ14}. Microsoft COCO contains 164k images in total where each image is labeled by at least 5 natural language captions. The performance is evaluated by human beings as well as some commonly used evaluation metrics (BLEU \cite{PapineniRWZ02}, Meteor \cite{denkowski2014meteor}, ROUGE-L \cite{lin2004rouge} and CIDEr-D \cite{VedantamZP15}). There are 17 teams participating the competition in total, and here we list the rankings of some top teams in Table \ref{table:app-cap-coco} including \cite{XuBKCCSZB15}.


\begin{table}[]
\centering
\caption{Rankings of some teams of the MS COCO Captioning Challenge. The rows are sorted by M1. For more details of the definitions of M1, and M2, one can refer to \footref{foot:app-cap-coco}. For team "Human", the captions are generated by human beings.}
\label{table:app-cap-coco}
\resizebox{\textwidth}{!}{%
\begin{tabular}{|l|l|l|l|l|l|l|l|l|l|}
\hline
\multirow{2}{*}{Model} & \multicolumn{2}{l|}{Human} & \multicolumn{7}{l|}{Other evaluation metrics}          \\ \cline{2-10}
                       & M1           & M2          & BLEU-1   & BLEU-2   & BLEU-3   & BLEU-4   & Meteor  & ROUGE-L   & CIDEr-D  \\ \hhline{|==========|}
Human                       & 1            & 1           & 6  & 12 & 12 & 13 & 3 & 11 & 6 \\ \hline
Google NIC                       & 2            & 3           & 2  & 2  & 2  & 2  & 1 & 1  & 1 \\ \hline
MSR \cite{FangGISDDGHMPZZ15}                       & 3            & 2           & 5  & 5  & 5  & 5  & 4 & 5  & 4 \\ \hline
\cite{XuBKCCSZB15}                       & 4            & 5           & 9  & 9  & 9  & 9  & 6 & 8  & 9 \\ \hline
MSR Captivator \cite{DevlinCFGDHZM15}                       & 5            & 4           & 1  & 1  & 1  & 1  & 2 & 2  & 2 \\ \hline
Berkeley LRCN \cite{DonahueHGRVDS15}                       & 6            & 6           & 10 & 7  & 8  & 8  & 7 & 6  & 8 \\ \hline
\end{tabular}
}
\end{table}

Except for the team "Human" and the system proposed by \cite{XuBKCCSZB15}, all other teams shown in Table \ref{table:app-cap-coco} use models without the attention mechanism. We can notice that the attention based model performs worse than NIC on both the human evaluations and the automatic evaluation metrics, which suggests that there are still many spaces for improvements. In addition, by comparing the human and the automatic evaluation metrics, the attention based RNN model performs worse in the automatic evaluation metrics than in the human evaluations. This may suggest that, on one side, the current automatic evaluation metrics are still not perfect, and on the other side, the attention based RNN model can generate captions which are more "natural" for human beings.

\begin{figure}[!ht]
  \centering
  \makebox[\textwidth]{\includegraphics[width=\textwidth]{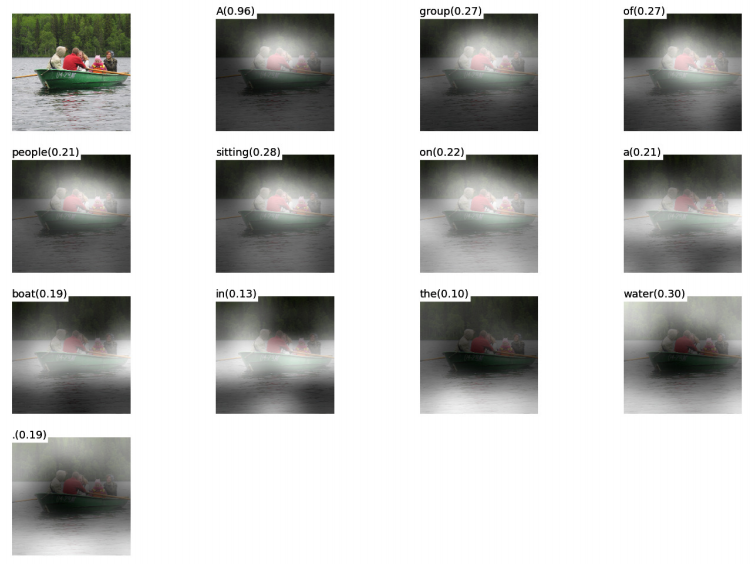}}
  \caption{A visual example of generating the caption with the item-wise soft attention based RNN model, where the white area indicates the position the model focuses on. The word at the top left corner is the word generated by the model. The figure is taken from \cite{XuBKCCSZB15}.}
  \label{fig:app-cap-saat-ex}
\end{figure}

In addition to the performance, another big advantage of the attention based RNN model is that it makes the visualization easier and much more intuitive as mentioned in Section \ref{sec:intro-adv}. Figure \ref{fig:app-cap-saat-ex} gives a visual example showing where the model attends in the decoding process. It is clear the attention mechanism indeed works, for example, when generating the word "people", the model focuses on the people, and when generating "water", the lake is attended.

The model in \cite{XuBKCCSZB15} uses fixed-size feature maps (14\texttimes14) to construct the code set $C$, and each feature map corresponds to a fixed-size patch of the input image. \cite{JinFCSZ15} claims that this design may harm the performance because some "meaningful scenes" may only occupy a small part of the image patch or cannot be cover by a single patch, where the meaningful scene indicates the objects/scenes correspond to the word is about to be predicted. \cite{JinFCSZ15} makes an improvement by firstly extract many object proposals \cite{UijlingsSGS13} from an input image, which potentially contain the meaningful scenes, as the input sequence. And then each proposal is used to calculate the code in $C$. However, \cite{JinFCSZ15} neither directly compares its performance to \cite{XuBKCCSZB15}, nor attends the MS COCO Captioning Challenge 2015. So here we cannot directly measure if and how the performance will be improved by using the object proposals to construct the input sequence. When analyzing in theory, we doubt if the improvements can really be obtained, because the currently popular object proposal methods only extract regions which probably contain some "objects". But a verb or an adjective in the ground truth image caption may not correspond to any explicit objects. Still, the problem proposed in \cite{JinFCSZ15} is very interesting and cannot be ignored: when the input of a RNN model is an image, it is not trivial to obtain a sequence of items from it. One straightforward solution is to use the location-wise attention models like the location-wise hard attention or the STN.

%% file: conclusion.tex
\section{Conclusion and future research}

In this survey, we discuss the attention based RNN model and its applications in some computer vision tasks. Attention is an intuitive methodology by giving different weights to different parts of the input which is widely used in the computer vision area. Recently with the development of the neural network, the attention mechanism is also embedded to the RNN model to further enhance its ability to solve sequence to sequence problem. We illustrate how the common RNN works, and gives detailed descriptions of four different attention mechanisms, and their pros and cons are also analyzed.

\begin{itemize}
  \item The item-wise attention based model requires that the input sequence contains explicit items. For some types of the input, like an image input, an additional step is needed to extract items from the input. Besides, the item-wise attention based RNN model needs to feed all items in the input sequence to the encoder, so on one side, it results in a slow training process, and on the other side, the testing efficiency is also not improved compared to the RNN without the attention module.
  \item The location-wise attention based model allows the input to be an entire feature map. At each time, the location-wise attention based model only focuses on a part of the input feature map, which gives efficiency improvements in both training and testing compared to the item-wise attention model.
  \item The soft attention module is differentiable with respect to its inputs, so the standard gradient ascent/decent can be used for optimization.
  \item The hard attention module is non-differentiable, and the objective is optimized by some techniques from the reinforcement learning.
\end{itemize}

At last, some applications in computer vision which apply the attention based RNN models are introduced. The first application is image classification and object detection which applies the location-wise hard attention mechanism. The second one is the image caption which uses both the item-wise soft and hard attention. The experimental results demonstrate that

\begin{itemize}
  \item Considering the performance, the attention based RNN model is better than, or at least comparable to, the model without the attention mechanism.
  \item The attention based RNN model makes the visualization more intuitive.
\end{itemize}

The location-wise soft attention model, i.e., the STN, is not introduced in the applications, because it is firstly proposed on the CNN and currently there are not enough applications of the STN based RNN model.

Since the attention based RNN model is a new topic, it has not been well addressed yet. As a result, there are many problems which either need theoretical analyses or practical solutions. Here we list a few of them:

\begin{itemize}
  \item A large majority of the current attention based RNN models are only applied on some small datasets. How to use the attention based RNN model to larger datasets, like the ImageNet dataset?
  \item Currently, most of the experiments are performed to compare the RNN models with/without the attention mechanism. More comprehensive comparisons of different attention mechanisms for different tasks are needed.
  \item In many applications the input sequences contain explicit items naturally, which makes them fit the item-wise attention model. For example, most of the applications in NLP treat the natural sentence as the input, where each item is a word in the sentence. Is it possible to convert these types of inputs to a feature map and apply the location-wise attention model?
  \item The current hard attention mechanism uses techniques from the reinforcement learning in optimization by maximizing an approximation of the sum of log-likelihood, which makes a sacrifice on the performance. Can the learning method be improved? \cite{BaGSF15} gives some insights but it is still an open question.
  \item How to extend the four attention mechanisms introduced in this survey? For example, an intuitive way of extension is to put multiple attention modules in parallel instead of only embedding one into the RNN model. Another direction is to further generalize an RNN and make it have a similar structure as the modern computer architecture \cite{weston2014memory, graves2014neural}, e.g. a multiple cache system shown in Figure \ref{fig:intro-cache}, where the attention module serves as a "scheduler" or an addressing module.
\end{itemize}

In summary, the attention based RNN model is a newly proposed model, and there are still many interesting questions remaining to be answered.